\documentclass[10pt,journal,compsoc]{IEEEtran}

\ifCLASSOPTIONcompsoc
  \usepackage[nocompress]{cite}
\else
  \usepackage{cite}
\fi

\usepackage{makecell}
\usepackage{wrapfig}
\usepackage{graphicx}
\usepackage[table]{xcolor}
\definecolor{LightCyan}{rgb}{0.88,1,1}
\definecolor{LightYellow}{rgb}{1,1,0.7}
\usepackage{verbatim}
\usepackage{amsmath}
\usepackage[colorlinks]{hyperref}
\usepackage[capitalize]{cleveref}

\usepackage[percent]{overpic}
\usepackage{pifont}
\usepackage{algorithm}
\usepackage{algpseudocode}
\usepackage{multirow}
\usepackage{float}
\usepackage{array}
\usepackage{tikz}
\usepackage{bm}
\usepackage{bbding}
\usepackage{pifont}
\usepackage{wasysym}
\usepackage{amssymb}
\usepackage{pifont}
\usepackage{tabularx}
\usepackage{booktabs}
\usepackage[export]{adjustbox}
\usepackage{enumitem}
\usepackage{arydshln}
\usepackage{subcaption}
\usepackage{adjustbox}
\usepackage{pdfpages}

\definecolor{lightgray}{gray}{0.9}
\definecolor{handcrafted}{gray}{0.1}
\definecolor{deep_learning}{RGB}{28, 0, 63}
\usepackage{xspace} 
\def\eg{\emph{e.g.}\xspace}

\Crefname{equation}{Eq.}{Eqs.}
\Crefname{figure}{Fig.}{Figs.}
\Crefname{tabular}{Tab.}{Tabs.}
\Crefname{section}{Sec.}{Secs.}

\usepackage{caption}
\definecolor{gold}{rgb}{1.0, 0.874, 0}
\definecolor{silver}{rgb}{0.77,0.77,0.77}
\definecolor{brown}{rgb}{0.95, 0.678, 0.4}

\definecolor{mylightgray}{RGB}{238,238,238} 
\colorlet{bgcolor}{mylightgray}

\definecolor{firstcolor}{HTML}{BDE6CD}
\definecolor{secondcolor}{HTML}{E2EEBC}
\definecolor{thirdcolor}{HTML}{FFF8C5}

\newcommand{\gold}[1]{\colorbox{firstcolor}{\textbf{#1}}}
\newcommand{\silver}[1]{\colorbox{secondcolor}{\textbf{#1}}}
\newcommand{\bronze}[1]{\colorbox{thirdcolor}{\textbf{#1}}}

\definecolor{maincategories}{HTML}{CDE1DB} 
\definecolor{salmon}{HTML}{EDF4F2} 

\usepackage{titlesec}

\titleformat{\paragraph}
{\normalfont\normalsize\itshape}{\theparagraph)}{1em}{}
\renewcommand{\theparagraph}{\alph{paragraph}}
\titlespacing*{\paragraph}
{0pt}{1.25ex plus 1ex minus .1ex}{1.ex plus .1ex}

\usepackage{tabulary}
\newcolumntype{I}{!{\vrule width 1pt}}
\newcolumntype{x}[1]{>{\centering\arraybackslash}p{#1pt}}
\newcolumntype{y}[1]{>{\raggedright\arraybackslash}p{#1pt}}
\newcolumntype{z}[1]{>{\raggedleft\arraybackslash}p{#1pt}}

\definecolor{higher}{HTML}{9FC4D9} 
\definecolor{lower}{HTML}{f9b0c7}  
\makeatletter
\newcommand{\thickhline}{%
    \noalign {\ifnum 0=`}\fi \hrule height 1pt
    \futurelet \reserved@a \@xhline
}
\makeatother

\title{A Survey on Deep Stereo Matching \\ in the Twenties}
\date{July 2022}

\author{Fabio Tosi \hspace{2cm} Luca Bartolomei \hspace{2cm} Matteo Poggi
\IEEEcompsocitemizethanks{\IEEEcompsocthanksitem F. Tosi, Luca Bartolomei and M. Poggi are with the Department of Computer Science and Enegineering, University of Bologna, Italy, 40136.
}}

\begin{document}

\IEEEtitleabstractindextext{%
\begin{abstract}

Stereo matching is close to hitting a half-century of history, yet witnessed a rapid evolution in the last decade thanks to deep learning. While previous surveys in the late 2010s covered the first stage of this revolution, the last five years of research brought further ground-breaking advancements to the field. This paper aims to fill this gap in a two-fold manner: first, we offer an in-depth examination of the latest developments in deep stereo matching, focusing on the pioneering architectural designs and groundbreaking paradigms that have redefined the field in the 2020s;
second, we present a thorough analysis of the critical challenges that have emerged alongside these advances, providing a comprehensive taxonomy of these issues and exploring the state-of-the-art techniques proposed to address them. By reviewing both the architectural innovations and the key challenges, we offer a holistic view of deep stereo matching and highlight the specific areas that require further investigation. To accompany this survey, we maintain a regularly updated project page that catalogs papers on deep stereo matching in our \href{https://github.com/fabiotosi92/Awesome-Deep-Stereo-Matching}{Awesome-Deep-Stereo-Matching} repository.

\end{abstract}

}
\maketitle

\section{Introduction}
\label{sec:introduction}

Stereo matching -- the task of estimating dense disparity maps from a pair of rectified images -- has been a fundamental problem in computer vision for nearly half a century, playing a crucial role in a wide range of applications, such as autonomous driving, robotics, and augmented reality.
After about twenty-five years of hand-designed stereo algorithms \cite{scharstein2002taxonomy}, the use of end-to-end deep neural networks has become the dominant paradigm in the late 2010s. 

Existing surveys \cite{poggi2021synergies,laga2020survey} have offered valuable insights on this rapid revolution, categorizing end-to-end architectures into 2D and 3D classes according to their cost-volume computation and optimization strategies, while also emphasizing the challenges remaining open.

Since then, however, the domain has progressed rapidly, with the emergence of novel methods and paradigms inspired by breakthroughs in other areas of deep learning. Transformer-based \cite{Li_2021_ICCV_STTR} and iterative refinement \cite{lipson2021raft} architectures are prime examples of how the field has evolved, demonstrating the potential for further improvements in accuracy and efficiency.

Despite the remarkable achievements, multiple challenges have emerged as deep stereo matching has advanced. One of the most critical issues, already highlighted in previous surveys, is the lack of generalization, particularly when facing domain shifts between synthetic and real data. Although synthetic datasets have been crucial for pre-training deep stereo networks, these models often perform poorly when applied to real scenes without fine-tuning.  Recognizing the importance of this issue, several techniques have been proposed in recent years to improve zero-shot generalization and to develop domain adaptation methods for seamless adaptation to unknown target domains.

Alongside the domain-shift problem, deep stereo matching has overcome several other limitations. The tendency of deep networks to over-smooth depth at object boundaries, leading to inaccurate 3D reconstructions, has been a persistent challenge.  The increasing diversity of camera setups and the need to process images with different resolutions has underscored the importance of developing flexible algorithms. The demand for high-resolution and highly detailed disparity estimation, combined with the necessity for real-time performance on resource-constrained devices, has further complicated the task. These challenges have led to the development of more robust, versatile, and efficient deep stereo models that effectively address these limitations and improve their practical application.

Furthermore, the integration of complementary imaging modalities, including depth sensors, non-visible spectrum cameras, event cameras, and gated cameras, has opened new avenues for enhancing the robustness and accuracy of stereo matching in challenging environments. 
By leveraging the strengths and complementary nature of these modalities, multimodal stereo matching techniques aim to extend the applicability and reliability of stereo vision to a broader range of real-world scenarios, overcoming the limitations of relying solely on visible-spectrum cameras.

Driven by these significant developments, this survey  provides a thorough and up-to-date review of the latest advancements in deep stereo matching that have emerged during the 2020s. We cover a wide range of topics explored since the last surveys \cite{poggi2021synergies,laga2020survey}, including novel architectures, multi-modal approaches, and cutting-edge techniques designed to address the aforementioned critical challenges. Our overview comprehensively analyzes over 100 distinct contributions to deep stereo matching, all of which have been presented at top conferences and published in prestigious journals.  Our goal is to offer a thorough overview of the state-of-the-art, discussing the progress made since previous surveys and highlighting the current trends and future directions in the field.

The remaining sections are organized as follows:

\begin{itemize}
    \item Section \ref{sec:architectures} covers the latest developments in deep stereo matching architectures, highlighting novel paradigms, computational efficiency techniques, and specialized multi-modality designs.
    \item Section \ref{sec:challenges} explores key challenges in deep stereo matching and investigates recent approaches to address them.
    \item Section \ref{sec:experiments} reports quantitative results on the main online benchmarks, both established and recently introduced, demonstrating the progress made in the field over the last few years. 
    \item Sections \ref{sec:discussion} and \ref{sec:conclusion} focus on potential future research directions and provide a summary of the survey. 
\end{itemize}

\ifx\arxiv\undefined

\else
For a more detailed overview of the background information, we recommend referring to the supplementary material. This additional resource provides detailed descriptions of the datasets commonly used in the context of deep stereo matching and the primary evaluation metrics employed in this domain. 
\fi
\ifx\supplementary\undefined
\subsection{Background}

To better understand the latest trends in the field, we introduce the fundamentals of deep stereo matching that have been driving advancements until the late 2010s.
In the interest of space, we recommend referring to the supplementary material for a detailed overview of the background information. This additional resource also provides detailed descriptions of the datasets commonly used in this context, as well as the primary evaluation metrics employed in this domain.
Finally, for a more comprehensive overview and detailed descriptions of the body of research that arose before 2020, readers can refer to existing surveys related to deep stereo matching, such as \cite{poggi2021synergies,laga2020survey, poggi2021confidence, poggi2017quantitative}.

\else

\begin{figure*}[t]
    \centering
    \begin{overpic}
    [clip,trim=0cm 23cm 50cm 0cm,width=\linewidth]{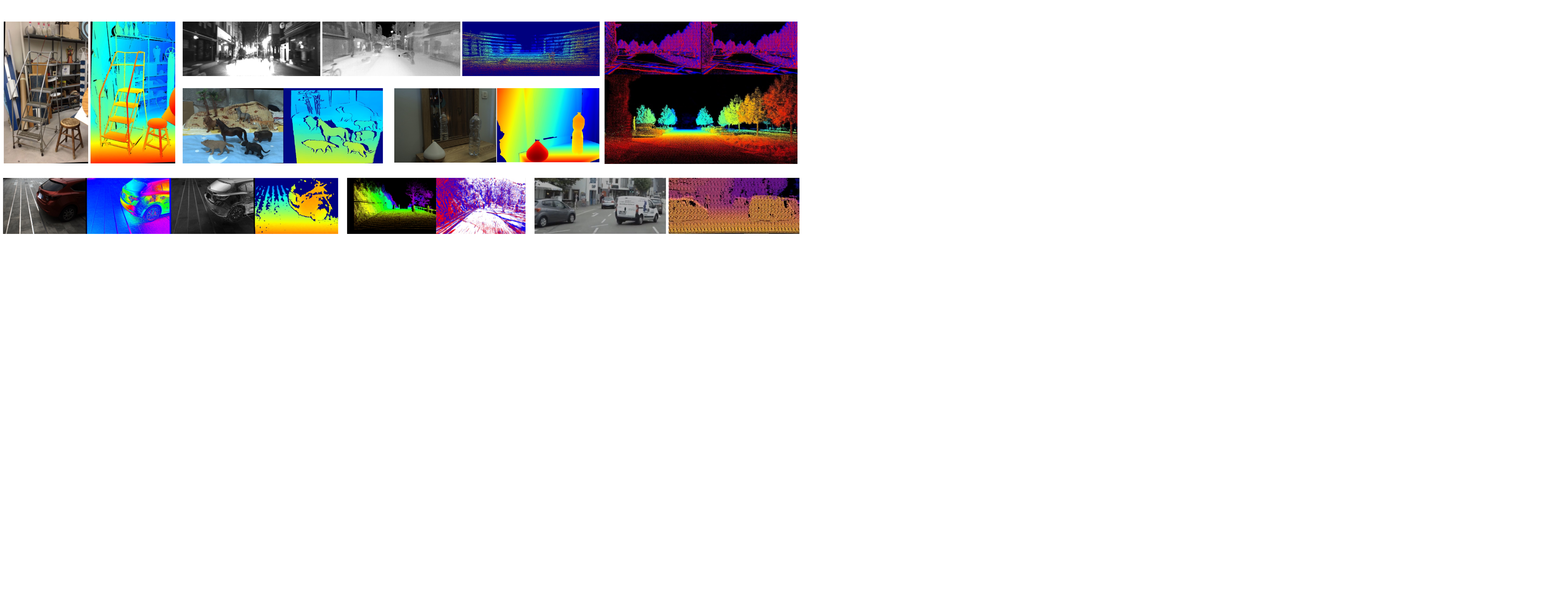}
    \put (48.5,20.8) {\scalebox{.65}{$MS^2$ \cite{Shin_2023_CVPR}}}
    \put (8,9.5) {\scalebox{.65}{Middlebury 2021 \cite{scharstein2014high}}}
    \put (33.5,9.5) {\scalebox{.65}{InStereo2K \cite{bao2020instereo2k}}}
    \put (60.7,9.5) {\scalebox{.65}{Booster \cite{ramirez2023booster}}}
    \put (86.5,9.5) {\scalebox{.65}{M3ED \cite{Chaney_2023_CVPR}}}
    \put (20,0.5) {\scalebox{.65}{RPS \cite{Tian_2023_ICCV}}}
    \put (53.5,0.5) {\scalebox{.65}{DSEC \cite{gehrig2021dsec}}}
    \put (82.5,0.5) {\scalebox{.65}{Gated Stereo \cite{walz2023gated}}}
    \end{overpic}\vspace{-0.3cm}
    \caption{\textbf{New stereo datasets in the 20s (Real-World).}}
    \label{fig:real_datasets}
\end{figure*}

\section{Background}
\label{sec:background}

We introduce the fundamentals of deep stereo matching that have been driving advancements until the late 2010s. 
For a more comprehensive overview and detailed descriptions of the body of research that arose before 2020, readers can refer to existing surveys related to stereo matching, such as \cite{poggi2021synergies,laga2020survey,poggi2017quantitative}.

\subsection{Theory}
\subsubsection{Learning for the Stereo Pipeline}
\label{sec:theory-stereo-pipeline}

In the early days, deep learning was studied for improving individual components of the traditional stereo pipeline \cite{scharstein2002taxonomy}. Specifically, CNN-based approaches were explored to learn more robust and discriminative matching cost functions, optimize the cost volume, and refine the final disparity map.

One of the primary areas of interest was the development of learned matching cost functions to replace hand-crafted ones, such as the sum of absolute differences (SAD) or the census transform (CT) \cite{zabih1994non}. These learned cost functions, typically implemented using Siamese CNN architectures, were trained to predict the similarity between image patches extracted from stereo pairs, resulting in more accurate and robust matching costs \cite{vzbontar2016stereo_MC-CNN, chen2015deep_Deep_Embed, luo2016efficient_Content_CNN}. The resulting cost volumes were then processed using conventional optimization techniques, such as SGM \cite{hirschmuller2007stereo}, and refined using traditional post-processing steps including bilateral and/or median filtering.

Some efforts were also made to improve the cost volume optimization and refinement stages using learning-based approaches. Several methods have been proposed to learn how to modulate the cost volume based on the reliability of matching costs \cite{park2015leveraging}, select highly confident pixels as constraints for optimization \cite{spyropoulos2014learning}, and adapt the aggregation step in SGM to reduce streaking artifacts \cite{poggi2016learning}. CNNs were also employed to refine the final disparity map \cite{shaked2017improved,gidaris2017detect, batsos2018recresnet,aleotti2021neural}, replacing conventional filtering techniques.

These early learning-based approaches demonstrated that deep learning could improve stereo matching by replacing certain hand-crafted components with learned methods, while still relying on the traditional pipeline structure.

The success of these methods in improving individual components of the stereo pipeline paved the way for the development of end-to-end deep stereo networks, which would come to dominate the field in the following years. By showing that learned components could outperform their hand-crafted counterparts, these early approaches laid the groundwork for the paradigm shift towards fully learnable models that directly estimate disparity maps from stereo image pairs.

\subsubsection{End-to-End Deep Stereo}
\label{sec:theory-end-to-end}

The advent of deep learning has revolutionized the field of stereo matching, enabling the development of end-to-end models that directly estimate disparity maps from stereo image pairs. These models have largely replaced traditional stereo matching pipelines, which typically consist of multiple hand-crafted steps \cite{scharstein2002taxonomy}. 
According to \cite{poggi2021synergies,laga2020survey}, end-to-end deep stereo networks directly estimating disparity maps from stereo image pairs can be broadly categorized into two main classes, based on their architecture: 2D networks and 3D networks.

Both 2D and 3D stereo networks begin by extracting deep features from the left and right input images using shared-weight CNNs. The key difference lies in how they construct and process the cost volume, which encodes the similarity between features at different disparity levels.

2D networks, such as DispNet \cite{mayer2016large}, usually build a 3D cost volume by computing the correlation between features at corresponding pixels across a range of disparities, encoding the similarity between patches centered on each pixel:

\begin{equation}
c(x_1, x_2) = \sum_{o\in[-k,k]\times[-k,k]} \langle f_1(x_1 + o),f_2(x_2 + o) \rangle
\end{equation}

where $f_1$ and $f_2$ are the features from the left and right images, and $k$ determines the neighborhood size. The resulting 3D cost volume is processed using 2D convolutions using an encoder-decoder design, inspired by the U-Net model \cite{ronneberger2015u}, to directly regress disparity values for each pixel. The use of plain 2D convolutions allows these networks to achieve real-time performance.

In contrast, 3D networks, first introduced by GC-Net \cite{kendall2017end_GC-NET} and later followed by more advanced architectures such as PSMNet \cite{chang2018pyramid_PSMNet} and GA-Net \cite{zhang2019ga}, construct a 4D cost volume by concatenating or computing the difference between features at all possible disparities. This 4D cost volume is then processed using 3D convolutions, which explicitly encode the geometry of the scene and capture the relationships between pixels at different disparity levels. The final disparity map is obtained using a fully differentiable soft argmin operation, which allows for sub-pixel disparity estimates:

\begin{equation}
\text{soft-argmin} = \sum_{d=0}^{D} d \cdot \sigma(-c_d)
\end{equation}

where $\sigma$ is the softmax operator applied along the disparity dimension $D$, and $c_d$ are the cost values at each disparity level $d$. First, the predicted costs $c_d$ from the cost volume are converted to a probability volume by taking the negative of each value. The probability volume is then normalized across the disparity dimension using the softmax operation, $\sigma(\cdot)$. Finally, the disparity is computed as the sum of each disparity $d$ weighted by its normalized probability. While 3D networks generally achieve higher accuracy than their 2D counterparts, they have significantly higher computational and memory requirements due to the use of 3D convolutions.

While 3D networks generally achieve higher accuracy than their 2D counterparts, they have significantly higher computational and memory requirements due to the additional dimension. Various techniques have been proposed to mitigate this computational burden, such as coarse-to-fine strategies \cite{khamis2018stereonet, khamis2018stereonet, wang2019anytime, yin2019hierarchical}, adaptive search space pruning \cite{duggal2019deeppruner}, and hierarchical architectures \cite{yang2019hierarchical_HSM}, and multi-scale feature extraction. 

Additionally, multi-task learning approaches have been explored to leverage the complementary nature of tasks like semantic segmentation \cite{yang2018segstereo} and edge detection \cite{song2019edgestereo}. For a more comprehensive overview and detailed descriptions of these methods, readers can refer to existing surveys, such as \cite{poggi2021synergies,laga2020survey}.

\fi
\section{Architectures}
\label{sec:architectures}

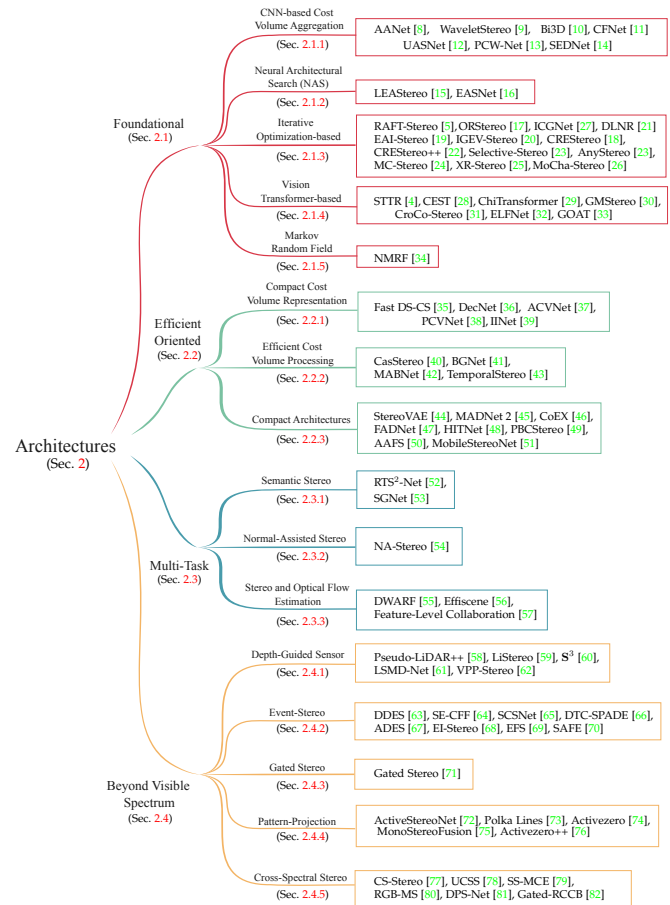
\begin{figure}[t]
  \centering
\includegraphics[width=0.48\textwidth]{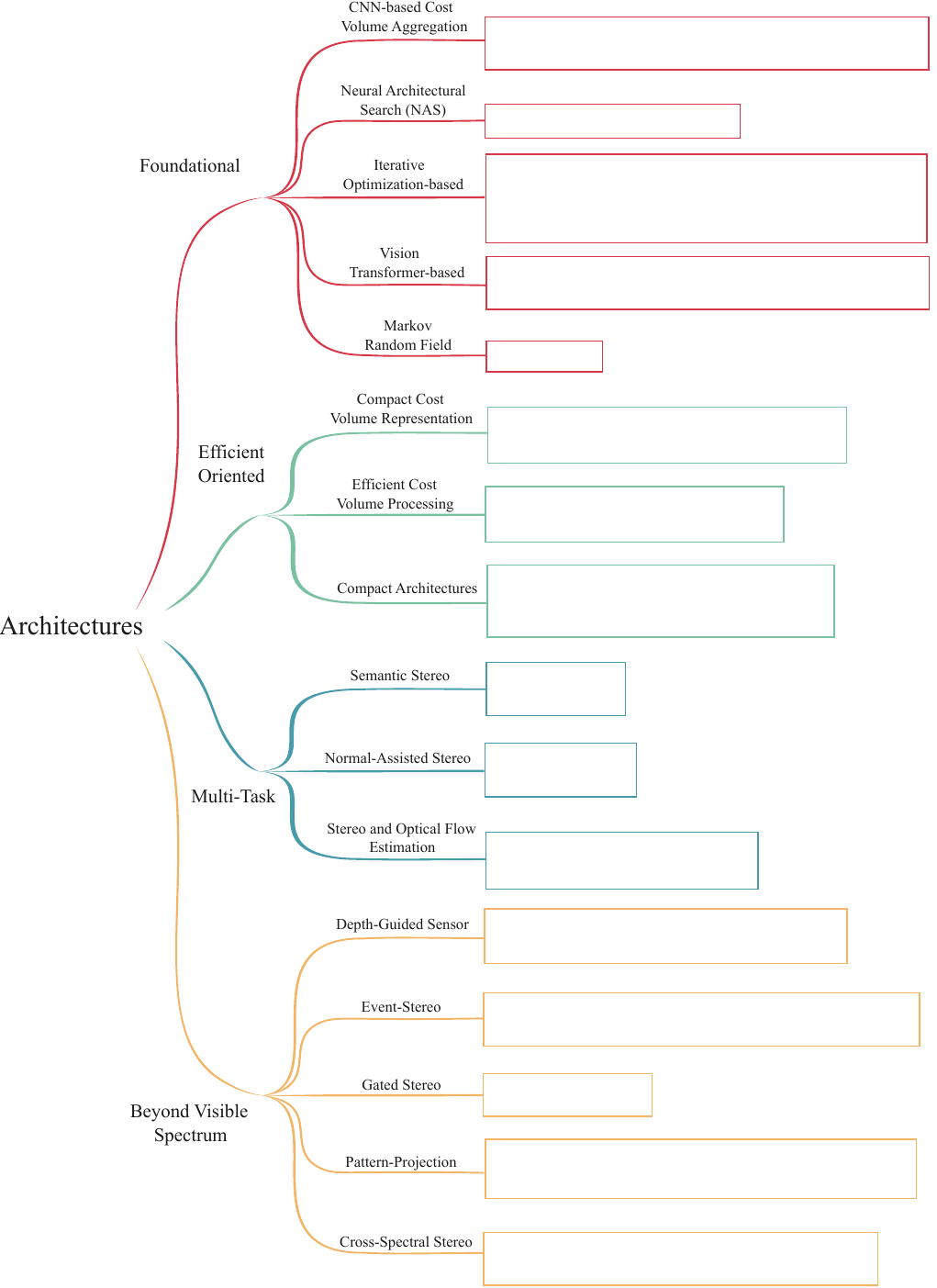}
\put (-237.,167) {\scalebox{.55}{(Sec. \ref{sec:architectures})}}
\put (-206.,290.5) {\scalebox{.45}{(Sec. \ref{sec:foundational-architectures})}}
\put (-206.,33) {\scalebox{.45}{(Sec. \ref{sec:beyond-visible-spectrum})}}
\put (-195.,208) {\scalebox{.45}{(Sec. \ref{sec:efficient-oriented})}}
\put (-195.,123.5) {\scalebox{.45}{(Sec. \ref{sec:multi-task-architectures})}}
\put (-150.,326) {\scalebox{.45}{(Sec. \ref{sec:cnn-based-cost-volume})}}
\put (-113.,331.5) {\scalebox{.45}{AANet \cite{xu2020aanet},}}
\put (-88.,331.5) {\scalebox{.45}{WaveletStereo \cite{Yang_2020_CVPR_WaveletStereo},}}
\put (-50.,331.5) {\scalebox{.45}{Bi3D \cite{Badki_2020_CVPR_Bi3D},}}
\put (-30.,331.5) {\scalebox{.45}{CFNet \cite{shen2021cfnet}}}
\put (-102.,325.5) {\scalebox{.45}{UASNet \cite{Mao_2021_ICCV_UASNet},}}
\put (-75.,325.5) {\scalebox{.45}{PCW-Net \cite{shen2022pcw},}}
\put (-46.5,325.5) {\scalebox{.45}{SEDNet \cite{chen2023learning}}}
\put (-150.,304) {\scalebox{.45}{(Sec. \ref{sec:nas-based})}}
\put (-113.,307.5) {\scalebox{.45}{LEAStereo \cite{cheng2020hierarchical_LEAStereo},}}
\put (-82.,307.5) {\scalebox{.45}{EASNet \cite{wang2022easnet}}}
\put (-113.,295.) {\scalebox{.45}{RAFT-Stereo \cite{lipson2021raft},}}
\put (-81.,295.) {\scalebox{.45}{ORStereo \cite{hu2021orstereo},}}
\put (-47.5,290.) {\scalebox{.45}{CREStereo \cite{li2022practical_CREStereo},}}
\put (-150.,283.5) {\scalebox{.45}{(Sec. \ref{sec:iterative_optimization-based})}}
\put (-113,290.) {\scalebox{.45}{EAI-Stereo \cite{zhao2022eai},}}
\put (-82,290.) {\scalebox{.45}{IGEV-Stereo \cite{xu2023iterative_IGEV-Stereo},}}
\put (-27,295.) {\scalebox{.45}{DLNR \cite{zhao2023high_DLNR}}}
\put (-113,285.) {\scalebox{.45}{CREStereo++ \cite{Jing_2023_ICCV_CREStereo++},}}
\put (-77,285.) {\scalebox{.45}{Selective-Stereo \cite{wang2024selective},}}
\put (-36,285.) {\scalebox{.45}{AnyStereo \cite{wang2024selective},}}
\put (-113,280.) {\scalebox{.45}{MC-Stereo \cite{Feng_2024_3DV_MC-Stereo},}}
\put (-82,280.) {\scalebox{.45}{XR-Stereo \cite{cheng2024stereo},}}
\put (-53,280.) {\scalebox{.45}{MoCha-Stereo \cite{mocha}}}
\put (-52.5,295.) {\scalebox{.45}{ICGNet \cite{gong2024learning},}}
\put (-113,266.5) {\scalebox{.45}{STTR \cite{Li_2021_ICCV_STTR},}}
\put (-94.5,266.5) {\scalebox{.45}{CEST \cite{guo2022context_CEST},}}
\put (-73.5,266.5) {\scalebox{.45}{ChiTransformer \cite{Su_2022_CVPR_Chitransformer},}}
\put (-33,266.5) {\scalebox{.45}{GMStereo \cite{xu2023unifying_GMStereo},}}
\put (-105,261.5) {\scalebox{.45}{CroCo-Stereo \cite{croco_v2},}}
\put (-150.,261) {\scalebox{.45}{(Sec. \ref{sec:vit-based})}}
\put (-69,261.5) {\scalebox{.45}{ELFNet \cite{lou2023elfnet},}}
\put (-43.5,261.5) {\scalebox{.45}{GOAT \cite{liu2024global_GOAT}}}
\put (-113,245) {\scalebox{.45}{NMRF \cite{guan2024neural_NMRF}}}
\put (-150.,242) {\scalebox{.45}{(Sec. \ref{sec:mrf-based})}}
\put (-113,226.5) {\scalebox{.45}{Fast DS-CS \cite{Yee_2020_WACV_FDCSC},}}
\put (-150.,221.5) {\scalebox{.45}{(Sec. \ref{sec:compact-cost-volume})}}
\put (-81,226.5) {\scalebox{.45}{DecNet \cite{Yao_2021_CVPR},}}
\put (-55,226.5) {\scalebox{.45}{ACVNet \cite{Xu_2022_CVPR_ACVNet},}}
\put (-95,221) {\scalebox{.45}{PCVNet \cite{zeng2023parameterized_PCVNet},}}
\put (-69,221) {\scalebox{.45}{IINet \cite{Li_Zhang_Su_Tao_2024_IINet}}}
\put (-113,206) {\scalebox{.45}{CasStereo \cite{pang2017cascade}, BGNet \cite{xu2021bilateral_BGNet},}}
\put (-113,201) {\scalebox{.45}{MABNet \cite{xing2020mabnet}, TemporalStereo \cite{Zhang2023TemporalStereo}}}
\put (-150.,199.5) {\scalebox{.45}{(Sec. \ref{sec:efficient-cost-volume})}}
\put (-113,185) {\scalebox{.45}{StereoVAE \cite{chang2023stereovae}, MADNet 2 \cite{Poggi_2024_CVPR_FedStereo}, CoEX \cite{bangunharcana2021correlate_CoEX}, }}
\put (-113,180) {\scalebox{.45}{FADNet \cite{wang2020fadnet}, HITNet \cite{Tankovich_2021_CVPR_HITNet}, PBCStereo \cite{Cai_2022_ACCV_PBCStereo},}}
 \put (-113,175) {\scalebox{.45}{AAFS \cite{Chang_2020_ACCV_AAFS}, MobileStereoNet \cite{Shamsafar_2022_WACV_MobileStereoNet}}}
 \put (-150.,176.5) {\scalebox{.45}{(Sec. \ref{sec:compact-architectures})}}
 \put (-150.,153.5) {\scalebox{.45}{(Sec. \ref{sec:semantic-stereo-matching})}}
\put (-113,160) {\scalebox{.45}{RTS$^2$-Net \cite{dovesi2020real}, }}
\put (-113,154) {\scalebox{.45}{SGNet \cite{chen2020sgnet}}}
 \put (-113,136) {\scalebox{.45}{NA-Stereo \cite{Kusupati_2020_CVPR} }}
  \put (-150.,132) {\scalebox{.45}{(Sec. \ref{sec:normal-assisted-stereo-matching})}}
 \put (-113,115) {\scalebox{.45}{DWARF \cite{aleotti2020learning},
 Effiscene \cite{jiao2021effiscene}, }}
 \put (-113,110) {\scalebox{.45}{Feature-Level Collaboration \cite{Chi_2021_CVPR}}}
   \put (-150.,107.5) {\scalebox{.45}{(Sec. \ref{sec:sceneflow})}}
 \put (-113,94) {\scalebox{.45}{Pseudo-LiDAR++ \cite{you2020pseudo}, LiStereo \cite{zhang2020listereo}, $\mathbf{S}^3$ \cite{huang2021s3}, }}
 \put (-113,89) {\scalebox{.45}{LSMD-Net \cite{Yin_2022_ACCV}, VPP-Stereo \cite{bartolomei2023active}}}
  \put (-150.,88) {\scalebox{.45}{(Sec. \ref{sec:depth-guided-sensor})}}
 \put (-113,72) {\scalebox{.45}{DDES \cite{tulyakov-et-al-2019}, SE-CFF \cite{Nam_2022_CVPR_CFF}, SCSNet \cite{cho2022selection_SCSNet}, DTC-SPADE \cite{Zhang_2022_CVPR_DTC_SPADE},}}
 \put (-113,67) {\scalebox{.45}{ADES \cite{Cho_2023_CVPR_ADES}, EI-Stereo \cite{mostafavi2021event_intensity}, EFS \cite{cho2022event}, SAFE \cite{Chen_2024_WACV_SAFE}}}
   \put (-150.,67) {\scalebox{.45}{(Sec. \ref{sec:event-stereo-matching})}}
 \put (-113,50) {\scalebox{.45}{Gated Stereo \cite{walz2023gated} }}
\put (-150.,45.5) {\scalebox{.45}{(Sec. \ref{sec:gated-stereo-matching})}}
 \put (-113,32) {\scalebox{.45}{ActiveStereoNet \cite{Zhang_2018_ECCV}, Polka Lines \cite{Baek_2021_CVPR}, Activezero \cite{liu2022activezero},}}
 \put (-113,28) {\scalebox{.45}{ MonoStereoFusion \cite{Xu_2022_CVPR}, Activezero++ \cite{chen2023activezero++} }}
 \put (-150.,26) {\scalebox{.45}{(Sec. \ref{sec:pattern-projection})}}
  \put (-113,9) {\scalebox{.45}{CS-Stereo \cite{zhi2018deep}, UCSS \cite{liang2019unsupervised}, SS-MCE \cite{walters2021there}, }}
  \put (-113,4) {\scalebox{.45}{RGB-MS \cite{tosi2022rgb}, DPS-Net \cite{Tian_2023_ICCV}, Gated-RCCB \cite{Brucker_2024_CVPR}}} 
 \put (-150.,4) {\scalebox{.45}{(Sec. \ref{sec:cross-spectral-stereo-matching})}}
  
  \caption{\textbf{A taxonomy of deep learning-based stereo matching architectures in the 2020s.} We categorize the reviewed methods based on their key designs and paradigms.}
  \label{fig:taxonomy}
\end{figure}

\subsection{Foundational Deep Stereo Architectures}
\label{sec:foundational-architectures}
\hypertarget{sec:architecture}{}This section categorizes and analyzes classical deep stereo models that emerged in recent years. These are grouped into five categories: CNN-based cost volume aggregation (Sec. \ref{sec:cnn-based-cost-volume}), Neural Architecture Search (NAS) for stereo matching (Sec. \ref{sec:nas-based}), iterative optimization-based (Sec. \ref{sec:iterative_optimization-based}), Vision Transformer-based (Sec. \ref{sec:vit-based}), and Markov random field-based architectures (Sec. \ref{sec:mrf-based}). These represent the primary trends and techniques that have significantly advanced the state-of-the-art of stereo depth estimation. The taxonomy depicted in Fig. \ref{fig:taxonomy} summarizes these categories.


\subsubsection{CNN-based Cost Volume Aggregation}
\label{sec:cnn-based-cost-volume}
Here, we cover recent deep stereo methods using established CNN-based cost volume aggregation techniques \cite{poggi2021synergies,laga2020survey}, 
which broadly fall into two categories: 2D and 3D architectures, distinguished by their strategy for encoding features and geometry. Both construct a cost volume from the left and right input images - 2D architectures typically employ a correlation layer on the extracted features, while 3D architectures concatenate or compute the feature difference over the full disparity range. This cost volume is then processed via plain 2D convolutions in 2D encoder-decoder architectures, or through 3D convolutions in 3D architectures that explicitly encode geometry. 

Among them, \hypertarget{AANet}{\textbf{AANet}} [\href{https://github.com/haofeixu/aanet}{Code}] \cite{xu2020aanet} aims to replace computationally expensive 3D convolutions while maintaining high accuracy by introducing a sparse points based intra-scale cost aggregation method using deformable convolutions to address the edge-fattening issue at disparity discontinuities, and an approximation of traditional cross-scale cost aggregation using neural network layers to handle large textureless regions. In contrast, \hypertarget{Bi3D}{\textbf{Bi3D}} [\href{https://github.com/NVlabs/Bi3D}{Code}]   \cite{Badki_2020_CVPR_Bi3D} formulates stereo matching as a collection of binary classification tasks, training a neural network to classify whether an object is closer or farther than a given depth plane, enabling selective depth estimation within a specific range of interest and offering a flexible trade-off between accuracy and computational efficiency.
\hypertarget{WaveletStereo}{\textbf{WaveletStereo}} \cite{Yang_2020_CVPR_WaveletStereo} takes a different approach by learning wavelet coefficients of the disparity map instead of directly estimating disparity. It comprises low-frequency and high-frequency sub-modules to handle smooth and detailed regions, respectively, and reconstructs the disparity map iteratively by mapping multi-resolution cost volumes to wavelet coefficients through convolutional networks and inverse wavelet transforms. \hypertarget{CFNet}{\textbf{CFNet}} [\href{https://github.com/gallenszl/CFNet}{Code}]  \cite{shen2021cfnet}, on the other hand, combines a fused cost volume representation, which fuses multiple low-resolution cost volumes to capture global and structural information, and a cascade cost volume representation, which adaptively adjusts the disparity search range at each stage using variance-based uncertainty estimation.
\hypertarget{UASNet}{\textbf{UASNet}} \cite{Mao_2021_ICCV_UASNet} constructs cascade 3D cost volumes with improved disparity range prediction and effective sampling by introducing an uncertainty distribution-guided range prediction (URP) module to handle ambiguities and an uncertainty-based disparity sampler (UDS) module that discretizes the per-pixel predicted disparity range based on the matching uncertainty. \hypertarget{PCW-Net}{\textbf{PCW-Net}} [\href{https://github.com/gallenszl/PCWNet}{Code}] \cite{shen2022pcw} employs a multi-scale cost volume fusion module to construct combination volumes on the upper levels of the pyramid and integrate them for initial disparity estimation, covering multi-scale receptive fields and extracting domain-invariant structural cues. It also introduces a warping volume-based disparity refinement module to narrow down the residue searching range from the initial disparity searching range to a fine-grained one.
\hypertarget{SEDNet}{\textbf{SEDNet}} [\href{https://github.com/lly00412/SEDNet}{Code}] \cite{chen2023learning} focuses on joint disparity and uncertainty estimation by extending the stereo backbone GWCNet \cite{guo2019group_GWCNet} with a lightweight uncertainty estimation subnetwork. It matches the distribution of disparity errors with the distribution of uncertainty estimates using a KL divergence-based loss function enabled by a differentiable histogramming scheme, ensuring that the uncertainty estimates follow the same distribution as the true errors of the disparity estimator.


\subsubsection{Neural Architecture Search for Stereo Matching}
\label{sec:nas-based}

Neural architecture search (NAS) aims to automate the design of neural architectures, alleviating the manual effort required by human experts. While NAS has achieved success in high-level vision tasks like classification and detection, applying it to the dense prediction problem of stereo matching is challenging. State-of-the-art deep stereo networks are computationally intensive, making the n\"aive application of traditional NAS methods prohibitively expensive due to the vast search space. To overcome this limitation, recent works have proposed tailored NAS frameworks that incorporate task-specific priors from the stereo matching pipeline. The key motivation is to enable efficient architecture search for this domain while maintaining high accuracy, without solely relying on manually designed architectures. 
\hypertarget{LEAStereo}{\textbf{LEAStereo}} [\href{https://github.com/XuelianCheng/LEAStereo}{Code}] \cite{cheng2020hierarchical_LEAStereo} is the first end-to-end hierarchical NAS framework for stereo that incorporates task-specific human knowledge. It consists of a 2D feature net for local image feature extraction and a 3D matching net for computing and aggregating matching costs from a 4D feature volume. The cell-level structure of these sub-networks is searched using a novel residual cell design and tailored candidate operations. At the network level, the search explores the arrangement of cells and the size of intermediate feature maps and volumes. This hierarchical approach, which embeds geometric priors, leads to accurate stereo matching models without relying on human-designed architectures.
Building upon LEAStereo, \hypertarget{EASNet}{\textbf{EASNet}} [\href{https://github.com/HKBU-HPML/EASNet}{Code}] \cite{wang2022easnet} addresses its limitations, such as high computational costs and limited scalability, by introducing a stereo network that can be deployed on devices with varying computing resources. The architecture search space covers depth, width, kernel size, and scale, allowing the network to adapt to different resource constraints. A progressive shrinking training strategy enables efficient optimization, allowing sub-networks with diverse configurations to be extracted without additional training while maintaining high accuracy.

\subsubsection{Iterative Optimization-based Architectures}
\label{sec:iterative_optimization-based}

Iterative optimization-based approaches have emerged as a promising paradigm for stereo matching, offering significant advantages over existing deep cost aggregation methods. Inspired by the success of iterative refinement in optical flow estimation, particularly the RAFT architecture \cite{teed2020raft}, these methods have been adapted to stereo matching with impressive results. By bypassing explicit cost aggregation and instead iteratively updating a disparity estimate using a high-resolution cost volume, these methods address key limitations of conventional approaches. 
Specifically, iterative stereo enables the direct use of high-resolution cost volumes without the computational burden of 3D convolutions, making it applicable to high-resolution images. Moreover, by avoiding a predefined disparity range, it can handle a wide range of disparities without sacrificing accuracy or efficiency. In fact, the recurrent nature of the update operator allows for a flexible trade-off between accuracy and speed through a variable number of iterations.

\begin{figure*}[t]
  \centering
  \begin{tabular}{@{}c@{}}
    \includegraphics[width=0.8\linewidth]{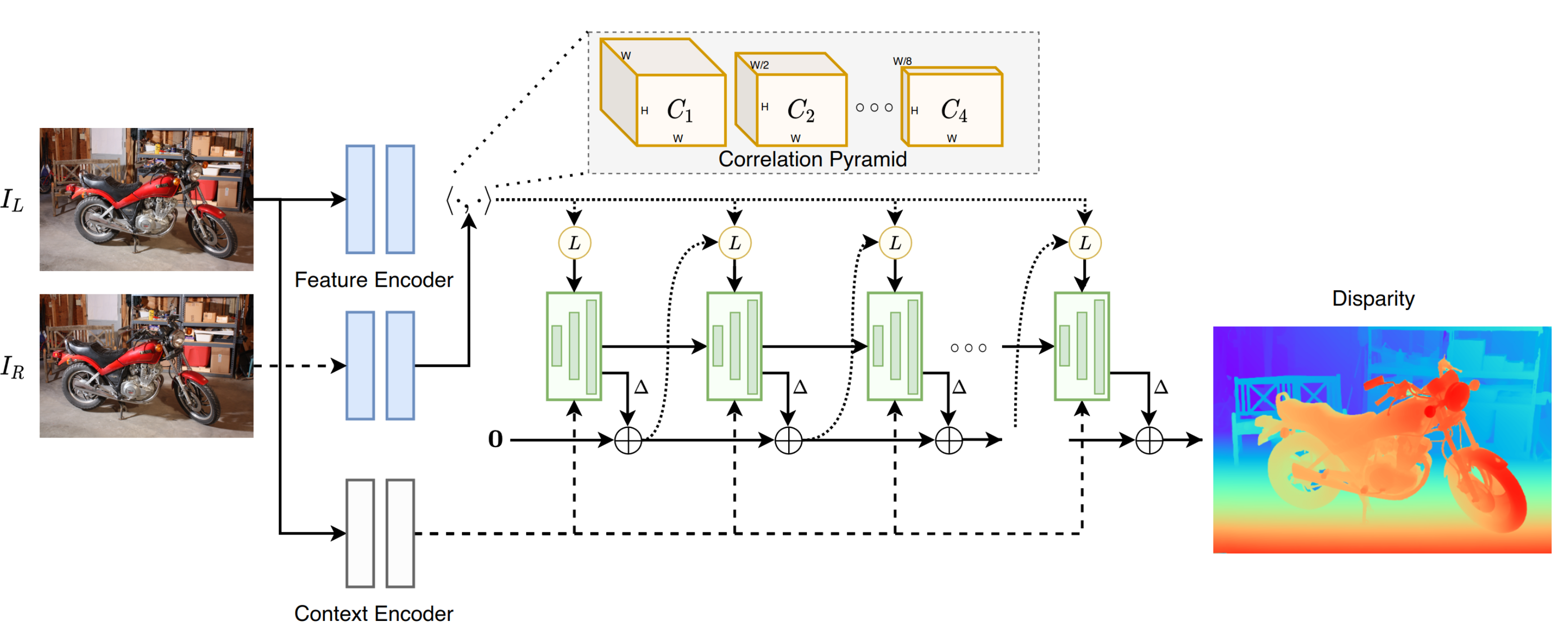}
  \end{tabular}
  \vspace{-0.7em}
  \caption{\textbf{RAFT-Stereo \cite{lipson2021raft} architecture.} It constructs a correlation pyramid from correlation features (blue) extracted from each image. A context encoder extracts ``context" image features (white) and an initial hidden state. The disparity field starts at zero. In each iteration, the GRU(s) (green) sample from the correlation pyramid using the current disparity estimate. Sampled correlation features, initial image features, and current hidden state(s) are processed by the GRU(s) to update the hidden state and disparity estimate. Picture from \cite{lipson2021raft}.}
  \label{fig:raft-stereo}
\end{figure*}

\textbf{RAFT-Stereo} [\href{https://github.com/princeton-vl/RAFT-Stereo}{Code}] \cite{lipson2021raft}\hypertarget{RAFT-Stereo}{, a deep stereo architecture inspired by the optical flow network RAFT \cite{teed2020raft}, has been a game-changer for recent stereo solutions. By introducing novel principles, illustrated in Fig. \ref{fig:raft-stereo}, such as lightweight cost volumes, and iterative refinement, RAFT-Stereo has paved the way for more efficient and effective processing of high-resolution stereo pairs. The architecture consists of three main components: a feature extractor, a correlation pyramid, and a Convolutional Gated Recurrent Units (ConvGRU)-based update operator. The feature extractor comprises a feature encoder, applied to both images to generate dense feature maps, and a context encoder, applied only to the left image to initialize the update operator and inject context features.
The correlation pyramid is constructed without predefining  the range of disparities by computing the visual similarity between feature vectors, restricting the computation to pixels sharing the same y-coordinate, resulting in a 3D cost volume:
\begin{equation}
\mathbf{C}_{ijk} = \sum_{h} \mathbf{f}_{ijh} \cdot \mathbf{g}_{ikh}, \quad \mathbf{C} \in \mathbb{R}^{H \times W \times W}
\end{equation}
Here, $\mathbf{C}$ represents the cost volume, $\mathbf{f}$ and $\mathbf{g}$ are the dense feature maps extracted from the left and right images, respectively. This lightweight cost volume is computed using a single matrix multiplication, making it computationally efficient.
The update operator iteratively retrieves local cost values from All-Pairs Correlations (APC) and updates the disparity field using multi-resolution ConvGRUs with cross-connections. The multi-level GRUs efficiently propagate information across the image, improving the global consistency of the disparity field. This iterative refinement process allows for a trade-off between accuracy and efficiency through early stopping. The final disparity is upsampled to the original resolution using a convex upsampling method.}
Building upon the principles introduced by RAFT-Stereo, several methods have proposed enhancements and adaptations to address specific challenges. Among them, \hypertarget{ORStereo}{\textbf{ORStereo}} \cite{hu2021orstereo} addresses the challenge of training stereo matching models on high-resolution images with large disparity ranges by proposing a two-phase approach. It estimates a down-sampled disparity map and occlusion mask in the first phase, followed by patch-wise refinement using a recurrent residual updater (RRU) and a normalized local refinement (NLR) module in the second phase. The RRU recurrently updates the disparity and occlusion predictions, while the NLR employs normalization techniques to generalize to unseen disparity ranges.
Building upon the principles of iterative refinement, \hypertarget{CREStereo}{\textbf{CREStereo}} [\href{https://github.com/megvii-research/CREStereo}{Code}] \cite{li2022practical_CREStereo} introduces a hierarchical network with recurrent refinement and an adaptive group correlation layer (AGCL) for robust matching. The AGCL computes correlations in local search windows, reducing memory and computation requirements, and incorporates deformable search windows and group-wise correlations. In contrast, \hypertarget{EAI-Stereo}{\textbf{EAI-Stereo}} [\href{https://github.com/smartadpole/EAI-Stereo}{Code}] \cite{zhao2022eai} focuses on error correction by incorporating an error-aware refinement module that combines left-right warping with learning-based upsampling, allowing the network to learn error correction capabilities.
\hypertarget{IGEV-Stereo}{\textbf{IGEV-Stereo}} [\href{https://github.com/gangweiX/IGEV}{Code}]\cite{xu2023iterative_IGEV-Stereo} takes a different approach by building a Combined Geometry Encoding Volume (CGEV) that encodes geometry, context information, and local matching details. It combines a Geometry Encoding Volume (GEV), obtained through lightweight 3D regularization of the cost volume, with All-Pairs Correlations to provide both non-local geometry and local matching details. \hypertarget{DLNR}{\textbf{DLNR}} [\href{https://github.com/David-Zhao-1997/High-frequency-Stereo-Matching-Network}{Code}] \cite{zhao2023high_DLNR}, on the other hand, focuses on preserving high-frequency information during the iterative process by incorporating a Decouple LSTM module, a Normalization Refinement module, and a multi-scale, multi-stage Channel-Attention Transformer feature extractor.
\hypertarget{CREStereo++}{\textbf{CREStereo++}} \cite{Jing_2023_ICCV_CREStereo++} extends CREStereo by introducing an uncertainty guided adaptive correlation (UGAC) module to enhance robustness and accuracy. It employs a content-aware warping layer with learnable offsets guided by variance-based uncertainty estimation, adjusting the sampling range when building the cost volume to capture more potential correspondences in ill-posed regions. \hypertarget{Selective-Stereo}{\textbf{Selective-Stereo}} [\href{https://github.com/Windsrain/Selective-Stereo}{Code}] \cite{wang2024selective} addresses the fixed receptive field limitation by proposing a Selective Recurrent Unit (SRU) with multiple GRU branches to adaptively capture and fuse multi-frequency information, and a Contextual Spatial Attention (CSA) module to guide the adaptive fusion based on image regions.
\hypertarget{Any-Stereo}{\textbf{Any-Stereo}} [\href{https://github.com/Zhaohuai-L/Any-Stereo}{Code}] \cite{Liang_Li_2024_Any-Stereo} focuses on maintaining the regularized disparity at a higher resolution by modeling the disparity as a continuous representation over 2D spatial coordinates using an Implicit Neighbor Mask Function (INMF). It proposes Intra-scale Similarity Unfolding (ISU) and Cross-scale Feature Alignment (CFA) strategies to complement missing information and details in the latent code. \hypertarget{XR-Stereo}{\textbf{XR-Stereo}} [\href{https://github.com/za-cheng/XR-Stereo}{Code}] \cite{cheng2024stereo}, instead, presents an architecture that leverages temporal information in video streams. It employs a RAFT-style scheme, where the key idea is to warp the previous frame's disparity and hidden state to the current frame using camera poses, serving as a warm-start for the GRU. 

\hypertarget{MC-Stereo}{\textbf{MC-Stereo}} [\href{https://github.com/MiaoJieF/MC-Stereo}{Code}] \cite{Feng_2024_3DV_MC-Stereo} addresses the limitations of existing approaches in handling multi-peak distribution and fixed search range by introducing a multi-peak lookup strategy and a cascade search range during the iterative process.
\hypertarget{MoCha-Stereo}{\textbf{MoCha-Stereo}}[\href{https://github.com/ZYangChen/MoCha-Stereo}{Code}] \cite{mocha} tackles the loss of geometric edge details by capturing repeated geometric contours across channels as motif channels that preserve common geometric structures. It employs a Motif Channel Attention (MCA) mechanism, a Motif Channel Correlation Volume (MCCV), and a Reconstruction Error Motif Penalty (REMP) module to recover lost geometric edge details through an iterative process.
Finally, \hypertarget{ICGNet}{\textbf{ICGNet}}[\href{https://github.com/DFSDDDDD1199/ICGNet}{Code}] \cite{gong2024learning} builds over existing architectures such as IGEV-Stereo and introduces additional intra-view and cross-view geometry constraints, by extracting keypoints with an off-the-shelf detector
and enforcing the image features extractor to predict the very same keypoints (intra-view), as well as by forcing the very correspondences between left and right image features as those occurring between left and right keypoints processed by an off-the-shelf matcher (cross-view).


\subsubsection{Vision Transformer-based Architectures}
\label{sec:vit-based}

In recent years, Vision Transformers (ViT) \cite{dosovitskiy2020image} have emerged as a promising alternative to CNNs for various computer vision tasks, including stereo matching. Transformers, originally developed for natural language processing \cite{vaswani2017attention}, have demonstrated remarkable performance in capturing long-range dependencies and global context information. These methods move away from traditional cost-volume construction and instead formulate stereo matching as a sequence-to-sequence problem, utilizing attention mechanisms to establish correspondences between pixels in the left and right images. Key components of vision transformer-based stereo matching include self-attention, cross-attention, and positional encoding schemes that provide spatial cues.

One of the pioneering works in this area is the STereo TRansformer (\hypertarget{STTR}{\textbf{STTR}}) [\href{https://github.com/mli0603/stereo-transformer}{Code}] \cite{Li_2021_ICCV_STTR}, which introduces a Transformer architecture that, unlike traditional cost-volume construction, relaxes the fixed disparity range constraint and alternates between self-attention along epipolar lines and cross-attention between left-right pairs. This approach allows the network to capture long-range pixel associations and resolve ambiguities in matching regions. Moreover, the architecture incorporates a relative positional encoding scheme to provide discriminative spatial cues and enforces a uniqueness constraint during matching, ensuring one-to-one correspondence between pixels across the stereo pair.

Building upon STTR, subsequent works have focused on enhancing the contextual information and refining the disparity estimation process. \hypertarget{CEST}{\textbf{CEST}} [\href{https://github.com/guoweiyu/Context-Enhanced-Stereo-Transformer}{Code}] \cite{guo2022context_CEST} introduces the Context Enhanced Path (CEP), a plug-in module that extracts long-range context information from low-resolution features. By following a coarse-to-fine approach and employing attention masking, optimal transport for uniqueness constraints, and post-processing modules like upsampling and context adjustment, CEST achieves improved disparity map refinement. Similarly, \hypertarget{ChiTransformer}{\textbf{ChiTransformer}} [\href{https://github.com/ISL-CV/ChiTransformer}{Code}] \cite{Su_2022_CVPR_Chitransformer} proposes a self-supervised method with an encoder-decoder transformer architecture, utilizing parallel ViT streams and depth cue rectification blocks to fuse binocular cues and produce depth estimates. While these methods focus on static stereo matching, \hypertarget{Dynamic-Stereo}{\textbf{Dynamic-Stereo}} [\href{https://dynamic-stereo.github.io/}{Code}] \cite{Karaev_2023_CVPR} tackles the challenge of temporally consistent disparity estimation from stereo videos. By introducing a hybrid encoder-decoder network that combines transformer-based attention across space, view, and time with an iterative refinement approach similar to RAFT, Dynamic-Stereo achieves good results in capturing dynamic scenarios.

Some works have also explored transformers for unifying various dense correspondence tasks. For instance, \hypertarget{GMStereo}{\textbf{GMStereo}} [\href{https://haofeixu.github.io/unimatch/}{Code}] \cite{xu2023unifying_GMStereo} presents a unified model for optical flow, rectified and unrectified stereo matching tasks, formulating them as a dense correspondence matching problem. By exploiting cross-attention in a Transformer-based model and employing parameter-free task-specific matching layers (2D, 1D, or another form) to obtain dense correspondences, GMStereo demonstrates the versatility and effectiveness of transformers in handling multiple related tasks.

Pre-training has also emerged as a promising approach to further improve the performance of ViT-based stereo matching models. \hypertarget{CroCo-Stereo}{\textbf{CroCo-Stereo}} [\href{https://github.com/naver/croco}{Code}] \cite{croco_v2} introduces a large-scale pre-training approach that leverages a cross-view completion pretext task to learn robust representations. The pre-trained model, CroCo v2, utilizes a ViT encoder and decoder architecture with self-attention and cross-attention mechanisms, and is then finetuned for stereo matching using a Dense Prediction Transformer (DPT) head, which directly processes the encoder and decoder features to predict the final disparity, in contrast to existing methods relying on cost volumes or iterative refinement.

Recognizing the complementary strengths of cost-volume-based and transformer-based approaches, \hypertarget{ELFNet}{\textbf{ELFNet}} [\href{https://github.com/jimmy19991222/ELFNet}{Code}] \cite{lou2023elfnet} proposes a novel framework that fuses the two paradigms. By employing heads in each branch to estimate aleatoric and epistemic uncertainties using deep evidential learning, and introducing a two-stage fusion strategy, ELFNet effectively integrates local and global information based on the estimated uncertainties.

The progressive refinement of disparity estimates in occluded regions is another crucial aspect addressed by ViT-based models. \hypertarget{GOAT}{\textbf{GOAT}} [\href{https://github.com/Magicboomliu/GOAT}{Code}] \cite{liu2024global_GOAT} introduces a parallel disparity and occlusion estimation module (PDO) and an iterative occlusion-aware global aggregation module (OGA). By exploiting restricted global spatial correlation within a focus scope guided by the occlusion mask, GOAT achieves robust disparity refinement in occluded regions.


\subsubsection{Markov Random Field-based Architectures}
\label{sec:mrf-based}

Markov Random Fields (MRFs) have been widely used in traditional stereo matching methods before the advent of deep learning, thanks to their ability to reduce matching ambiguities in challenging regions by enforcing spatial coherence and smoothness constraints on the disparity map. However, traditional MRF models rely on hand-crafted potential functions and message passing procedures, which often lead to suboptimal results. Recently, a new line of research has emerged that aims to leverage the power of deep learning to create fully data-driven MRF models for stereo. The first attempt in this direction is the Neural Markov Random Field (\hypertarget{NMRF}{\textbf{NMRF}}) model [\href{https://github.com/aeolusguan/NMRF}{Code}] \cite{guan2024neural_NMRF}. Specifically, NMRF introduces a novel fully data-driven approach that consists of a local feature CNN, a Disparity Proposal Network (DPN), a Neural MRF inference stage, and a disparity refinement stage. The local feature CNN extracts multi-level features from the stereo pair, while the DPN prunes the search space by identifying the top k disparity modals for each pixel through neural message passing. The inference stage performs coarse-level disparity estimation using a probabilistic graphical model with self and neighbor edges, where the potential functions and message passing are learned through neural networks. Finally, the disparity refinement stage further improves the coarse estimates. The model is trained using a combination of proposal loss, initialization loss, and disparity loss, which measures the discrepancy between candidate labels, cost volume, and the ground truth disparity. By learning complex pixel relationships and potential functions directly from data while retaining the graph inductive bias of MRFs, NMRF opens up new possibilities for combining the strengths of both traditional MRFs and deep learning in stereo matching.


\subsection{Efficiency-Oriented Architectures}
\label{sec:efficient-oriented}

\hypertarget{sec:efficiency}{While deep learning has enabled dramatic accuracy improvements over traditional stereo methodologies by training convolutional neural networks to learn powerful feature representations and cost volume filtering for better matching, many deep stereo networks are too computationally intensive for real-time operation, especially on embedded or mobile devices with strict power, memory, and computational constraints.  In such cases, there is an important need to develop efficient deep neural architectures that can provide the accuracy benefits of learned approaches while operating at real-time speeds with low latency on low-power edge devices.} 


\subsubsection{Compact Cost Volume Representations}
\label{sec:compact-cost-volume}
Efficient stereo matching often requires compact cost volume representations to reduce memory and computational demands. Various approaches have been proposed to extract essential matching information while suppressing redundant data, enabling faster processing without significant accuracy degradation. 

Among these frameworks, \hypertarget{Fast DS-CS}{\textbf{Fast DS-CS}} [\href{https://github.com/ayanc/fdscs}{Code}] \cite{Yee_2020_WACV_FDCSC}, unlike most existing stereo networks using expensive learned matching costs and 3D convolutions, uses traditional efficient matching costs to construct initial cost volumes and learns a mapping to convert the cost information at each pixel into a low-dimensional ``cost signature" feature vector, which is then processed using an encoder-decoder network with 2D convolutions. In contrast, \hypertarget{DecNet}{\textbf{DecNet}} [\href{https://github.com/YaoChengTang/DecNet}{Code}] \cite{Yao_2021_CVPR} proposes a decomposition model that reduces the computational cost growth as resolution increases by decomposing stereo matching into a dense matching at very low resolution and a series of sparse matchings at higher resolutions, enabling logarithmic complexity growth. \hypertarget{ACVNet}{\textbf{ACVNet}} [\href{https://github.com/gangweiX/ACVNet}{Code}] \cite{Xu_2022_CVPR_ACVNet} introduces an attention-based filtering approach, leveraging correlation clues and a multi-level adaptive patch matching mechanism to generate the Attention Concatenation Volume (ACV). \hypertarget{PCVNet}{\textbf{PCVNet}} [\href{https://github.com/jiaxiZeng/Parameterized-Cost-Volume-for-Stereo-Matching}{Code}] \cite{zeng2023parameterized_PCVNet}, instead, proposes a parameterized cost volume representation that encodes the entire disparity space using a multi-Gaussian distribution, allowing for a global view of the disparity space and progressively focusing on local regions for fine-grained matching. \hypertarget{IINet}{\textbf{IINet}} \cite{Li_Zhang_Su_Tao_2024_IINet} aims to address the redundancy in explicit 3D cost volumes by proposing a compact 2D implicit network, incorporating intra-image context information, Fast Multi-scale Score Volume (FMSV), Confidence Based Filtering (CBF), and Intra-Inter Fusing (IIF) network with Residual Context-aware Upsamplers (RCUs) for enhanced information transmission. These methods employ different strategies, such as cost signatures, dense-sparse decomposition, attention-based filtering, parameterized representations, and implicit networks, to achieve compact yet informative cost volume representations.


\subsubsection{Efficient Cost Volume Processing}
\label{sec:efficient-cost-volume}
Processing high-dimensional cost volumes can be a major bottleneck in real-time stereo matching. To address this, various techniques have been developed to efficiently process cost volumes, such as cascaded architectures, edge-aware upsampling, and multi-scale feature extraction. 

Accordingly, \hypertarget{CasStereo}{\textbf{CasStereo}} [\href{https://github.com/alibaba/cascade-stereo}{Code}] \cite{Gu_2020_CVPR_CasStereo} addresses this by proposing a cascade cost volume formulation, constructing a cost volume using a feature pyramid and iteratively narrowing down the depth range at each stage, recovering the output in a coarse-to-fine manner. In contrast, \hypertarget{BGNet}{\textbf{BGNet}} [\href{https://github.com/3DCVdeveloper/BGNet}{Code}] \cite{xu2021bilateral_BGNet} focuses on edge-preserving cost volume upsampling using a learned bilateral grid, enabling efficient upsampling while preserving sharp edges and allowing computationally expensive operations to be performed at lower resolutions without compromising accuracy. \hypertarget{MABNet}{\textbf{MABNet}} [\href{https://github.com/JumpXing/MABNet}{Code}] \cite{xing2020mabnet} introduces the Multibranch Adjustable Bottleneck (MAB) module for multi-level feature extraction, with the 2D MAB module having three branches with different dilation rates and the 3D MAB module factorizing 3D convolutions into disparity-wise and spatial convolutions, aiming for high accuracy with a small model size. \hypertarget{TemporalStereo}{\textbf{TemporalStereo}} [\href{https://github.com/youmi-zym/TemporalStereo}{Code}] \cite{Zhang2023TemporalStereo}, instead, employs a coarse-to-fine architecture with sparse cost volumes, enriching them with multi-level context, statistical fusion for robust, global cost aggregation, and an adaptive shifting strategy that adjusts the disparity candidates based on the aggregated costs. Moreover, it can seamlessly operate in both single stereo pair and temporal modes, leveraging past information to boost matching accuracy with a negligible runtime increase.

\subsubsection{Compact Architectures}
\label{sec:compact-architectures}
Computing the cost volume is the primary contributor to
computational complexity in end-to-end networks, but not the only one. Thus, designing an architecture that is compact in any part can further decrease complexity.

%
For instance, \hypertarget{StereoVAE}{\textbf{StereoVAE}}\cite{chang2023stereovae} \hypertarget{StereoVAE}{proposes a computationally efficient and lightweight framework combining a traditional matching algorithm and a variational autoencoder (VAE) based neural network. Initially, the former generates a coarse disparity map from a low-resolution stereo image pair, leveraging its low computational complexity. Subsequently, a tiny VAE-based network upscales and refines the coarse disparity map, taking the coarse map and the left image as inputs.} 

%
\textbf{MobileStereoNet} [\href{https://github.com/cogsys-tuebingen/mobilestereonet}{Code}] \cite{Shamsafar_2022_WACV_MobileStereoNet} \hypertarget{MobileStereoNet}{proposes a 3D and a 2D lightweight stereo models to pursue efficiency. The former builds upon GwcNet-g \cite{guo2019group_GWCNet}, while the latter uses 2D encoder-decoder to process a 3D cost volume, which is constructed efficiently using a new "Interlacing Cost Volume" module. Both models replace various components with 2D and 3D versions of MobileNet-V1 and MobileNet-V2 convolution blocks to further reduce complexity.}

Bringing lightweight design to the extreme, \hypertarget{PBCStereo}{\textbf{PBCStereo}} \cite{Cai_2022_ACCV_PBCStereo} is a fully binarized deep stereo network. To mitigate accuracy degradation due to quantization, PBCStereo introduces the Interpolation+Binary Convolution (IBC) module to replace binary deconvolutions and fuse shallow/deep features, and proposes the Binary Input Layer (BIL) coding method to binarize inputs while preserving pixel precision. PBCStereo incorporates IBC modules for any of feature extraction, cost aggregation, and disparity refinement stages. 
\hypertarget{FADNet}{\textbf{FADNet}} [\href{https://github.com/HKBU-HPML/FADNet}{Code}] \cite{wang2020fadnet} focuses on efficient 2D-based correlation layers with stacked residual blocks. 
It employs a DispNetC-based backbone extensively reformed with residual blocks, point-wise correlation layers, and multi-scale residual learning, adopting a coarse-to-fine training strategy with a loss weight scheduling technique.
\hypertarget{HITNet}{\textbf{HITNet}} [\href{https://github.com/google-research/google-research/tree/master/hitnet}{Code}] \cite{Tankovich_2021_CVPR_HITNet} avoids building a full 3D cost volume and uses fast multi-resolution initialization. 
The disparity map is represented as planar tiles with learnable feature descriptors at multiple resolutions, and the network iteratively refines disparity hypotheses via differentiable 2D geometric propagation and warping, reasoning about slanted plane hypotheses for accurate warping and upsampling in coarse-to-fine manner.
\hypertarget{CoEX}{\textbf{CoEX}} [\href{https://github.com/antabangun/coex}{Code}] \cite{bangunharcana2021correlate_CoEX}, instead, pursues efficiency and accuracy through the Guided Cost volume Excitation (GCE) module, which utilizes extracted image features from the MobileNetV2 \cite{sandler2018mobilenetv2} backbone to excite relevant channels in the cost volume, improving feature extraction without significant computational overhead. CoEX also introduces top-k selection before soft-argmin disparity regression, computing the final disparity estimate using only the top-k matching cost values.
\hypertarget{AAFS}{\textbf{AAFS}} [\href{https://github.com/JiaRenChang/RealtimeStereo}{Code}] \cite{Chang_2020_ACCV_AAFS} combines an efficient backbone made of depthwise separable convolutions, an attention-aware feature aggregation (ACCV) module, and a cascaded 3D CNN architecture to achieve real-time stereo on edge devices. The ACCV module adaptively aggregates information from different scales of the feature maps, enhancing their representational capacity, while the cascaded 3D CNN architecture regularizes cost volumes at multiple in coarse-to-fine manner.
Lastly, \textbf{MADNet 2} [\href{https://github.com/mattpoggi/fedstereo}{Code}] \cite{Poggi_2024_CVPR_FedStereo} revises MADNet \cite{Tonioni_2019_CVPR} with the all-pairs correlation module from RAFT-Stereo, while removing its original context network. Combined with refined augmentation techniques, these modifications enable MADNet 2 to significantly outperform its predecessor while preserving computational efficiency.


\subsection{Multi-task Deep Architectures}
\label{sec:multi-task-architectures}
In this section, we focus on models combining stereo matching with other dense tasks within a multi-task framework.

\subsubsection{Semantic Stereo Matching}
\label{sec:semantic-stereo-matching}
Integrating semantic information into stereo matching has the potential to significantly improve the performance of depth estimation algorithms. By exploiting high-level semantic cues, these methods can better handle challenging scenarios such as textureless regions, occlusions, and depth discontinuities. 
Despite the many attempts \cite{zhan2019dsnet, yang2018segstereo, jiang2019sense, wu2019semantic} reported in previous surveys \cite{poggi2021synergies}, only few further works have emerged in the 20s, specifically targeting real-time stereo matching with semantic guidance. 
Among them, \hypertarget{RTS2Net}{$\mathbf{RTS}^2\mathbf{Net}$}\cite{dovesi2020real} proposes a lightweight, real-time architecture for joint semantic segmentation and disparity estimation. Key aspects include a multi-stage, coarse-to-fine pyramidal decoder design for trading accuracy and speed, a shared encoder extracting common features for both tasks, separate semantic and disparity decoders operating on the shared features, and a synergy refinement module that exploits the complementarity between semantics and depth. 
\hypertarget{SGNet}{\textbf{SGNet}}\cite{chen2020sgnet}, instead, built upon PSMNet \cite{chang2018pyramid_PSMNet}, introduces three modules to embed semantic constraints: a confidence module computing the consistency between disparity and semantic features, a residual module optimizing the initial disparity map based on semantic categories, and a loss module supervising disparity smoothness based on semantic boundaries and regions.

\subsubsection{Normal-Assisted Stereo Matching}
\label{sec:normal-assisted-stereo-matching}
These methods aim to exploit the geometric constraints between depth and surface normals to regularize the depth estimation process and enhance the overall performance. One notable example of this approach is HITNet \cite{Tankovich_2021_CVPR_HITNet}, which achieves high accuracy by geometrically reasoning about disparities and inferring slanted plane hypotheses. Building upon a similar concept, \hypertarget{NA-Stereo}{\textbf{NA-Stereo}}\cite{Kusupati_2020_CVPR} was proposed for generic multi-view stereo that can also be applied to the classic binocular stereo matching task. The key idea is to incorporate a normal estimation network (NNet) into the depth estimation framework, jointly optimizing for both depth and normals through a consistency loss. The method constructs a 3D cost volume by plane sweeping and accumulates multi-view image information, which is then regularized by NNet that predicts surface normals from the accumulated features. 


\subsubsection{Joint Stereo And Optical Flow} 
\label{sec:sceneflow}

This family of frameworks processes stereo videos by jointly estimating three elements for each pair of consecutive left images: the disparity map for that image pair, the optical flow between the two left images, and optionally, the change in disparity between them.  Predicting these three cues together allows for estimating the 3D scene flow of the observed scene \cite{vedula1999three}.
It is worth mentioning that, among the literature, several works directly estimate 3D scene flow starting from point clouds or pre-computed disparity maps \cite{teed2021raft3d,liu2022camliflow}. Here, we survey only those approaches estimating disparity maps as well.
Among these, \hypertarget{DWARF}{\textbf{DWARF}} [\href{https://github.com/FilippoAleotti/DWARF-Tensorflow}{Code}]\cite{aleotti2020learning} proposed a compact coarse-to-fine architecture estimating disparity, flow, and disparity change at multiple scales, by building compact correlation volumes exploiting features warping across the scales. Specifically, a 3D correlation layer is deployed across the correlation scores computed between two consecutive stereo pairs.
On the contrary, \hypertarget{Effiscene}{\textbf{Effiscene}}\cite{jiao2021effiscene} tackles scene flow estimation in an unsupervised manner, by decomposing the task into predicting stereo depth, optical flow, camera poses, and motion segmentation. The four ingredients combined together allow for estimating the 3D scene flow of the scene in the absence of any annotation.
Finally, 
\hypertarget{Feature-Level Collaboration}{\textbf{Feature-Level Collaboration}}\cite{Chi_2021_CVPR} follows the same track and proposes an architecture for joint estimation of stereo depth and optical flow, followed by a pose decoder estimating the camera ego-motion from the previous two predictions.


\subsection{Beyond Visible Spectrum Deep Stereo Matching}
\label{sec:beyond-visible-spectrum}
In recent years, there has been a significant increase in the use of modalities beyond conventional color cameras for various tasks. This section examines frameworks that perform stereo estimation by combining color imagery with data from external depth sensors or different types of cameras, or by operating entirely outside the visible RGB spectrum.

\subsubsection{Depth-Guided Sensor Stereo Matching}
\label{sec:depth-guided-sensor}
Guided stereo matching aims to enhance the accuracy and robustness of stereo networks by leveraging sparse depth cues from external sensors, such as LiDAR. These depth hints, independent of visual appearance, help to overcome limitations in challenging scenarios like textureless regions 
and mitigate the impact of domain shift. 

\hypertarget{Pseudo-LiDAR++}{\textbf{Pseudo-LiDAR++}}\cite{you2020pseudo} adapts a stereo network to directly estimate depth by constructing a depth cost volume and refines the estimates using a graph-based depth correction algorithm that propagates sparse LiDAR measurements. In contrast, \hypertarget{LiStereo}{\textbf{LiStereo}}\cite{zhang2020listereo} employs a two-branch architecture, with a color image branch for stereo feature extraction and a LiDAR branch for processing sparse depth maps, which are then fused to output dense depth maps.

The Sparse Signal Superdensity \hypertarget{$\mathbf{S}^3$}{(\textbf{$\mathbf{S}^3$)}}\cite{huang2021s3} framework, instead, addresses the challenges of low density and imbalanced distribution of sparse depth cues by expanding the depth estimates around sparse signals based on the RGB image and controlling their influence through confidence weighting. \hypertarget{LSMD-Net}{\textbf{LSMD-Net}}\cite{Yin_2022_ACCV} fuses LiDAR and stereo information using a dual-branch disparity predictor, where the outputs from these branches are fused at each pixel using a mixture density module that models the estimated disparity at each pixel as a probability distribution using Laplacian distributions. The final disparity map is then determined by the expectation of the more confident branch.

Differently from the previous strategies, \hypertarget{VPP-Stereo}{\textbf{VPP-Stereo}} [\href{https://github.com/bartn8/vppstereo/}{Code}] \cite{bartolomei2023active} 
uses depth measurements from an active sensor to hallucinate patterns onto the stereo images, thus simplifying visual correspondence. Various patterning strategies are proposed to improve performance. The hallucinated stereo pair is processed by any stereo matching algorithm (either learned or hand-crafted) to obtain dense depth estimations. Lastly, \hypertarget{SDG-Depth}{\textbf{SDG-Depth}} [\href{https://github.com/SJTU-ViSYS/SDG-Depth}{Code}] \cite{li2024stereo} incorporates a sparse LiDAR deformable propagation module to generate a Semi-Dense hint Guidance map and a confidence map, which guide cost aggregation in stereo matching. Additionally, SDG-Depth introduces a learned disparity-depth conversion module to reduce triangulation errors. 


\subsubsection{Event-Camera-Based Stereo Matching} 
\label{sec:event-stereo-matching}

Event cameras \cite{event_survey}, also known as neuromorphic cameras, are peculiar sensors designed to mitigate the shortcomings of conventional imaging devices, such as limited dynamic range, sensor noise, and motion blur caused by rapid movements. In contrast to color cameras, which capture frames at fixed intervals, event cameras operate asynchronously. 
As will be discussed, the majority of these methods adapt successful models from traditional stereo literature \cite{poggi2021synergies} to this novel paradigm, primarily by developing appropriate techniques for handling asynchronous event streams. 

Specifically, given a timestamp $t_d$ at which disparity map estimation is desired, events are usually sampled backward in time from the stream, either based on a time interval (SBT) or a maximum number of events (SBN), and processed to obtain tensors ready to be processed by standard CNNs.

The pioneering effort dates back to the 2010s: \hypertarget{DDES}{\textbf{DDES}} [\href{https://github.com/tlkvstepan/event_stereo_ICCV2019}{Code}] \cite{tulyakov-et-al-2019},
\hypertarget{DDES}{ 
inspired by the previous successes in the classical stereo literature \cite{kendall2017end_GC-NET}. 
Events are organized in queues 
and processed by temporal fully connected layers, implemented as MLPs to aggregate the information encoded along time by the events, and used to build a 4D cost volume, optimized with a 3D UNet inspired by GC-Net \cite{kendall2017end_GC-NET}. }

%
More recent works delved into strategies for codifying the events streams: \textbf{SE-CFF} [\href{https://github.com/yonseivnl/se-cff}{Code}] \cite{Nam_2022_CVPR_CFF} \hypertarget{SE-CFF}{
aims at solving 
the loss of meaningful events due to overriding, by proposing a ConcentrationNet for aggregating the events streams into a compact representation preserving fine details. Furthermore, 
SE-CFF distills knowledge from future events during training to compensate for the missing information in past events. SE-CFF builds a correlation-based cost volume and processes it with dilated 2D convolutions \cite{xu2020aanet} and a 2D UNet to get the final prediction.}
%
%
Given the many events sampled from streams, \textbf{SCSNet}\cite{cho2022selection_SCSNet} \hypertarget{SCSNet}{
implements a differentiable mechanism for retaining only those relevant for estimating disparity at the desired timeframe. Furthermore, a Neighbor Cross Similarity Feature (NCSF) extraction module is designed to encode the similarity across the different modalities -- images and events. SCSNet uses group-wise correlation layers \cite{guo2019group_GWCNet} and 3D convolutions in its backbone.
} 
%
%
In contrast, \textbf{DTC-SPADE}\cite{Zhang_2022_CVPR_DTC_SPADE} \hypertarget{DTC-SPADE}{samples events according to SBT and stores them in voxel grids, that are processed by Discrete-Time Convolution (DTC) modules -- recurrent operators designed to embed data in a variety of temporal scales.
Furthermore, spatially adaptive denormalization modules (SPADE \cite{park2019semantic}) are used to complement the sparse voxel grid embeddings with edge information from the original raw events. 
Following DDES \cite{tulyakov-et-al-2019}, DTC-SPADE deploys a GC-Net-like backbone \cite{kendall2017end_GC-NET} to predict a dense disparity map.}
%
%
Peculiarly, \textbf{ADES}\cite{Cho_2023_CVPR_ADES} \hypertarget{ADES}{focuses on domain adaptation of an event-stereo model, either based on AANet \cite{xu2020aanet} or PSMNet \cite{chang2018pyramid_PSMNet}, trained on the pseudo-events frame generated by EventGAN \cite{eventgan} from color images. During adaptation on real events streams, these are used to generate pseudo-color images \cite{Rebecq19pami} and compute a smudge-aware self-supervised loss.
Furthermore, a Motion-invariant Consistency module is designed to perturb events and simulate slower or faster motion and then impose a consistency loss between the disparity maps obtained from perturbed and original inputs.}


Eventually, some frameworks emerged to process both color and events. \textbf{EI-Stereo} [\href{https://github.com/yonseivnl/se-cff}{Code}] \cite{mostafavi2021event_intensity} has been proposed to match pairs of event and color images to exploit the best of the two worlds. \hypertarget{EI-Stereo}{
An event-intensity recycling network iteratively processes each image with a single channel of the corresponding event stack -- from the most recent to the oldest -- while keeping a hidden state propagated through the channels.
Then, an AANet-like architecture \cite{xu2020aanet} is deployed for estimating the disparity map.}
On the same track, \textbf{EFS}\cite{cho2022event} \hypertarget{EFS}{deploys distinct feature extractors for images and events. The features are fused across the modalities and used to build a correlation-based, multi-scale cost volume, that passes through a 3D aggregation network to estimate a dense disparity map. An additional cost volume between event features is built at training time, and processed to obtain a sparse disparity map.}

Lastly, \textbf{SAFE}\cite{Chen_2024_WACV_SAFE} \hypertarget{SAFE}{estimates disparity from asymmetric inputs -- a color image and an event stream -- by dividing the task into three sub-problems: i) asymmetric color-event matching, ii) color Structure-from-Motion (SfM), and iii) event SfM. Two feature extractors process images and events, and then three cost volumes are built, i) between image and events, ii) time-adjacent images, and iii) time-adjacent events, and combined in a single volume, to be processed by a fusion module deploying 3D ConvLSTMs.}


\subsubsection{Gated Stereo Matching}
\label{sec:gated-stereo-matching}
Adverse weather conditions such as fog and snow can significantly affect the performance of conventional color cameras. Gated cameras \cite{andersson2006long} are designed to be robust against these conditions. During acquisition, an illuminator emits light to flood a specific range of the scene in front of it. Multiple acquisitions are performed by illuminating predetermined distances (slices) -- each one not interfering with the others -- and then accumulated in a single frame. \textbf{Gated Stereo}\cite{walz2023gated} \hypertarget{Gated Stereo}{
utilizes a synchronized wide-baseline gated stereo camera and deploys a network made of monocular and stereo branches, exploiting both time-of-flight intensity cues and multi-view cues, respectively. These branches are combined through a fusion network to produce the final depth map. 
The framework is trained in a semi-supervised manner, with a novel ambient-aware and illuminator-aware self-supervised loss on gated images.
}


\subsubsection{Pattern Projection-Based}
\label{sec:pattern-projection}
Pattern projection-based approaches for active stereo vision have gained significant attention in recent years due to their ability to enhance depth estimation accuracy and robustness in challenging scenarios. These methods employ a projector to cast a structured light pattern onto the scene, providing additional cues for stereo matching. By exploiting the known pattern geometry and the correspondence between the projected pattern and the captured images, pattern projection-based techniques can effectively handle textureless regions, and other ambiguities that often pose difficulties for traditional passive matching algorithms. The projected patterns are typically designed to have high spatial frequency and good uniqueness properties to facilitate accurate and efficient stereo matching. 

In the 10s, \hypertarget{ActiveStereoNet}{\textbf{ActiveStereoNet}} \cite{Zhang_2018_ECCV} appeared as the pioneering work in end-to-end learning for active stereo systems, introducing a novel reconstruction loss based on local contrast normalization (LCN) to address the challenges of illumination variations and textureless regions. Furthermore, ActiveStereoNet proposes an invalidation network to explicitly handle occlusions, which is trained end-to-end without ground-truth data. Building upon this foundation, In the 20s, \hypertarget{Polka Lines}{\textbf{Polka Lines}} \cite{Baek_2021_CVPR} takes a step further by jointly learning the structured illumination pattern and the reconstruction algorithm, parameterized by a diffractive optical element (DOE) and a neural network, respectively. This joint optimization approach enables the learned "Polka Lines" patterns to be tailored to the reconstruction network, achieving high-quality depth estimates across various conditions.
\hypertarget{Activezero}{\textbf{Activezero}} \cite{liu2022activezero} addresses the lack of real-world depth annotations by proposing a mixed-domain learning framework that combines supervised learning on synthetic data with self-supervised learning on real data. It introduces a novel temporal IR reprojection loss that is more robust to noise and textureless patches and invariant to illumination changes. 
Despite the similarities with Polka Lines, it focuses on improving generalization. 
\hypertarget{MonoStereoFusion}{\textbf{MonoStereoFusion}} \cite{Xu_2022_CVPR} takes a different path by integrating a monocular structured light subsystem and a binocular stereo subsystem to leverage their complementary advantages. The monocular subsystem handles textureless regions, while the binocular subsystem is used for distant objects and outdoor scenes.  In contrast to other methods, MonoStereoFusion introduces an IR camera without a narrow-band filter, allowing it to receive both visible and IR light simultaneously.
Lastly, \hypertarget{ActiveZero++}{\textbf{ActiveZero++}} \cite{chen2023activezero++} extends the ActiveZero framework by incorporating an illumination-invariant feature matching module and a confidence-based depth completion module. The illumination-invariant feature matching module uses a self-attention mechanism to learn robust features that are insensitive to illumination variations. The confidence-based depth completion module uses the confidence from the stereo network to identify and improve erroneous areas in depth prediction through depth-normal consistency.


\subsubsection{Cross-Spectral Stereo Networks}
\label{sec:cross-spectral-stereo-matching}
Cross-spectral stereo matching has emerged as a promising approach for estimating depth by finding correspondences between images captured in different spectral bands, such as visible (RGB) and near-infrared (NIR), short-wave infrared (SWIR), or thermal infrared (TIR). 
Indeed, in many real-world scenarios, RGB stereo matching often struggles to find reliable correspondences -- e.g., in low-light conditions, fog, etc. By leveraging the distinct appearance characteristics of materials and objects across different spectral bands, cross-spectral stereo aims to overcome these limitations. 
This comes with unique challenges, such as the significant appearance differences between images from different spectral bands or the scarcity of annotated, cross-spectral datasets.

Before the 20s, \textbf{CS-Stereo} \cite{zhi2018deep} processed RGB and NIR images without depth supervision. It simultaneously estimates disparity and translates the RGB image to a pseudo-NIR image using a disparity prediction network and a spectral translation network, incorporating a material recognition network to handle unreliable matching regions. Similarly, \textbf{UCSS} \cite{liang2019unsupervised} employed a spectral translation network based on F-cycleGAN to minimize appearance variations between cross-spectral images and a stereo matching network to estimate disparity, enabling iterative end-to-end unsupervised learning. \textbf{SS-MCE} \cite{walters2021there} takes a different approach, estimating dense flow fields between images of different spectra without requiring ground truth. It employs a dual-spectrum siamese-like structure with two flow estimation modules, each estimating the flow field to align one image with the other and then warp it back, demonstrating its versatility across different spectral combinations. \textbf{RGB-MS} \cite{tosi2022rgb} focuses on registering RGB and MS images with different resolutions using a self-supervised deep architecture consisting of coarse and fine-grained sub-modules, employing a proxy label distillation strategy for supervision. \textbf{DPS-Net} \cite{Tian_2023_ICCV}, instead, is an end-to-end network for polarimetric stereo depth estimation that leverages multi-domain similarity and geometric constraints, constructing RGB and polarization correlation volumes, introducing an iso-depth cost to handle polarization ambiguities, and employing a cascaded dual-GRU architecture to recurrently update the disparity. Finally, \textbf{Gated-RCCB} \cite{Brucker_2024_CVPR} is an approach that fuses high-resolution RCCB and gated NIR cameras to estimate dense, accurate depth maps. It exploits active and passive signals across visible and NIR spectra, proposing a cross-spectral stereo network with a fusion module to effectively integrate features from both modalities.

\section{Challenges and Solutions}
\label{sec:challenges}

\begin{figure}[t]
  \centering
\includegraphics[width=0.50\textwidth]{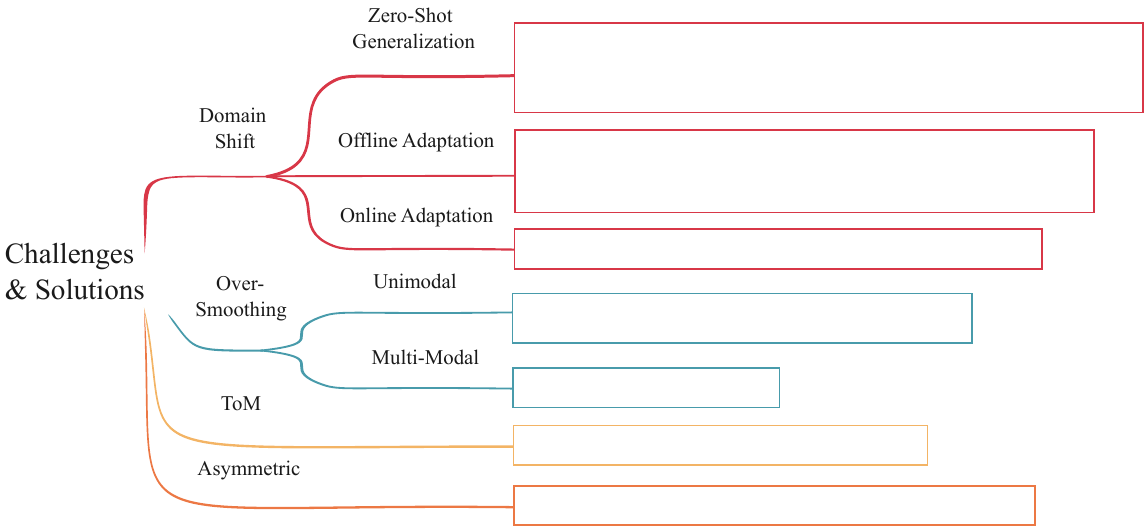}
\put (-215.,81) {\scalebox{.45}{(Sec. \ref{sec:generalization})}}
\put (-250.,45) {\scalebox{.55}{(Sec. \ref{sec:challenges})}}
\put (-215.,43) {\scalebox{.45}{(Sec. \ref{sec:over-smoothing})}}
\put (-215.,22) {\scalebox{.45}{(Sec. \ref{sec:tom})}}
\put (-215.,7) {\scalebox{.45}{(Sec. \ref{sec:asymmetric})}}
\put (-175.,103.7) {\scalebox{.45}{(Sec. \ref{sec:zero-shot})}}
\put (-175.,81) {\scalebox{.45}{(Sec. \ref{sec:offline-adaptation})}}
\put (-175.,64.5) {\scalebox{.45}{(Sec. \ref{sec:online-adaptation})}}
\put (-175.,50) {\scalebox{.45}{(Sec. \ref{sec:unimodal})}}
\put (-175.,33) {\scalebox{.45}{(Sec. \ref{sec:multimodal})}}
\put (-140,108) {\scalebox{.45}{DSMNet \cite{zhang2019domaininvariant_DSM}, FCStereo \cite{Zhang_2022_CVPR_FCStereo}, GraftNet \cite{Liu_2022_CVPR_GraftNet}, ITSA \cite{Chuah_2022_CVPR_ITSA}, HVT \cite{Chang_2023_CVPR}}} 
\put (-140,102) {\scalebox{.45}{MRL-Stereo \cite{Rao_2023_CVPR}, MS-Nets \cite{cai2020matchingspace}, ARStereo \cite{cheng2022revisiting_AdversariallyRobustStereo}, NDR \cite{aleotti2021neural,tosi2024neural}}} 
\put (-140,96) {\scalebox{.45}{EVHS \cite{Pilzer_2023_WACV_EVHS}, LSSI \cite{watson2020learning_LSSI}, NS-Stereo \cite{Tosi_2023_CVPR}, DKT-Stereo \cite{zhang2024robust_DKT}}} 
\put (-140,84) {\scalebox{.45}{Flow2Stereo \cite{Liu_2020_CVPR_Flow2Stereo}, Reversing-Stereo \cite{aleotti2020reversing}, Revealing-Stereo \cite{Chen_2021_ICCV},}}
\put (-140,79) {\scalebox{.45}{MultiscopicVision \cite{yuan2021stereo}, PASMNet \cite{shen2023digging}, StereoGAN \cite{Liu_2020_CVPR_StereoGAN}}}
\put (-140,74) {\scalebox{.45}{AdaStereo \cite{Song_2021_CVPR_AdaStereo}, UCFNet \cite{shen2023digging}, TiO-Depth \cite{Zhang_2022_CVPR}, RAG \cite{Zhang_2022_CVPR}}}
\put (-140,62) {\scalebox{.45}{AoHNet \cite{wang2020faster}, MAD++ \cite{poggi2021continual}, PointFix \cite{kim2022pointfix}, FedStereo \cite{Poggi_2024_CVPR_FedStereo}}}
\put (-140,46.5) {\scalebox{.45}{SM-CDE \cite{chen2019over_smoothing}, AcfNet \cite{zhang2020adaptive_AcfNet}, CDN \cite{garg2020wasserstein}, LaC \cite{liu2022local_LaC}}}
\put (-140,30) {\scalebox{.45}{SMD-Nets \cite{Tosi2021CVPR_SMD}, ADL \cite{Xu_CVPR_2024_ADL}}}
\put (-140,17.5) {\scalebox{.45}{DDF \cite{chai2020deep}, TA-Stereo \cite{wu2023transparent}, Depth4ToM \cite{costanzino2023learning}}}
\put (-140,3.5) {\scalebox{.45}{VI-Stereo \cite{liu2020visually}, NDR \cite{aleotti2021neural,tosi2024neural}, DA-AS \cite{Chen_2022_CVPR}, SASS \cite{song2023unsupervised}}}
  \caption{\textbf{A taxonomy of the main challenges (and solutions) in deep stereo matching.} For each, we highlight the key problem areas and novel techniques developed.}
  \label{fig:challenges}
\end{figure}
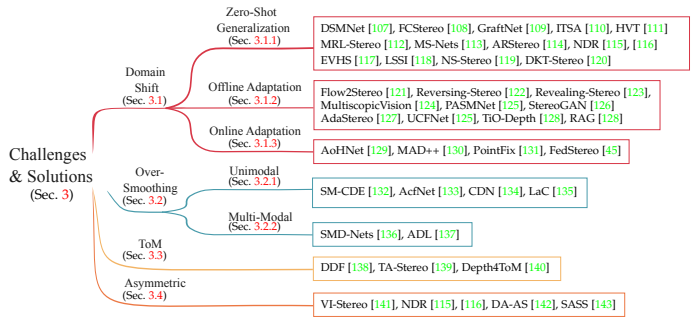

The development of more accurate stereo architectures in the 2020s allowed researchers to tackle many challenges that were unresolved in the late 2010s \cite{poggi2021synergies}. As illustrated in Fig. \ref{fig:challenges}, these challenges can be categorized into four main groups, discussed in the following sections.

\subsection{Domain Shift}
\label{sec:generalization}

Existing deep stereo networks achieve remarkable performance when trained and tested on data from the same domain. However, when these models are deployed to the target scenarios, they often encounter a significant drop in performance due to the domain shift between the training and testing data. This shift can be attributed to: 

\begin{itemize}
    \item Variations in color, illumination, contrast, and texture.
    \item Differences in stereo camera setups, such as baseline distance, focal length, and sensor properties. 
    \item Variations in the disparity range due to different depths of the scenes (\eg, indoor vs. outdoor environments).
\end{itemize}
These differences can make stereo networks learn domain-specific features or shortcuts that do not transfer well to novel scenarios. To address this challenge, several techniques have been explored lately. 

\subsubsection{Zero-Shot Generalization}
\label{sec:zero-shot}
\begin{figure*}[h]
\centering
\renewcommand{\tabcolsep}{1pt}
\begin{tabular}{ccccc}

& \multicolumn{2}{c}{Trained on SceneFlow \cite{butler2012naturalistic}, with GT} & \multicolumn{2}{c}{Trained on Real-World data \cite{Tosi_2023_CVPR}, without GT} \\
\includegraphics[width=0.18\textwidth]{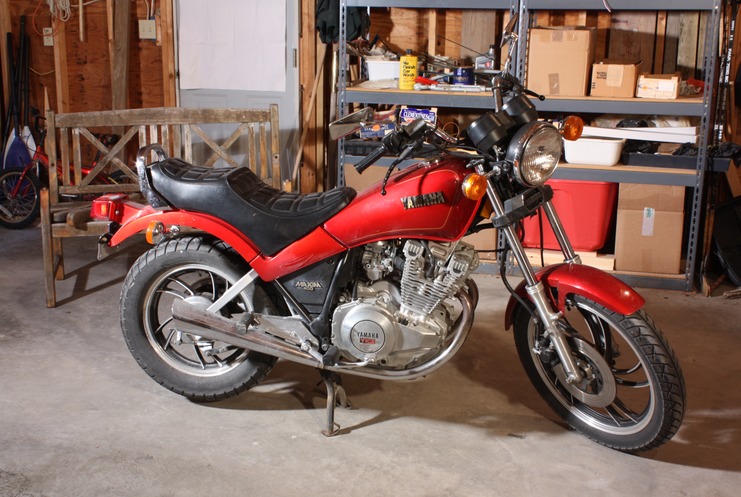} &
\includegraphics[width=0.18\textwidth]{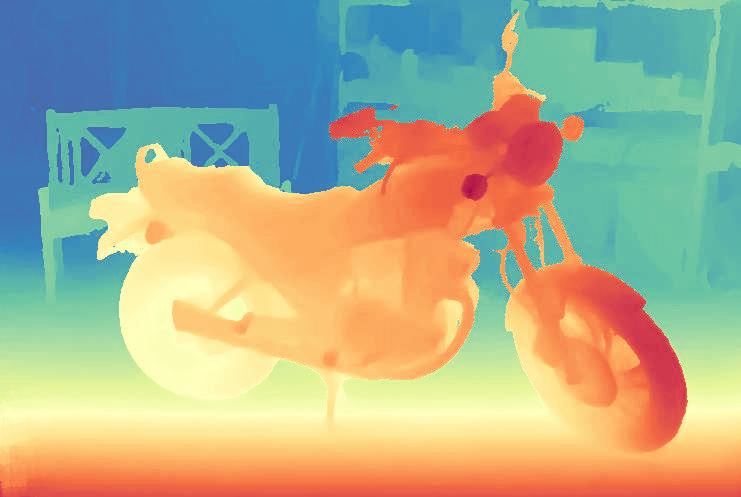} & 
\includegraphics[width=0.18\textwidth]{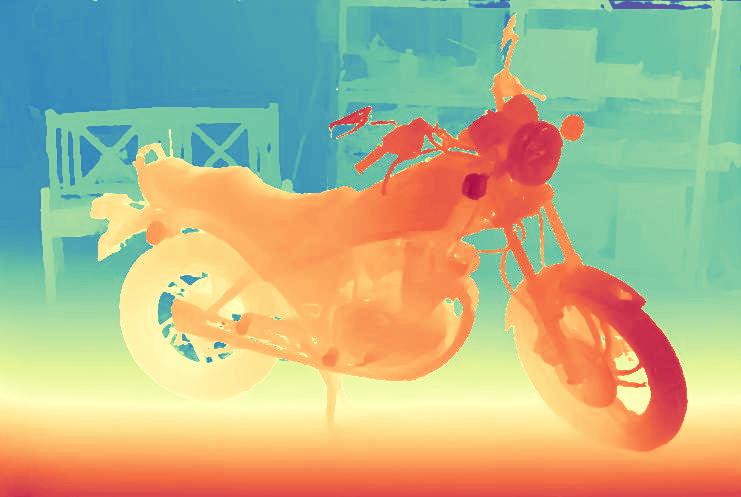} & 
\includegraphics[width=0.18\textwidth]{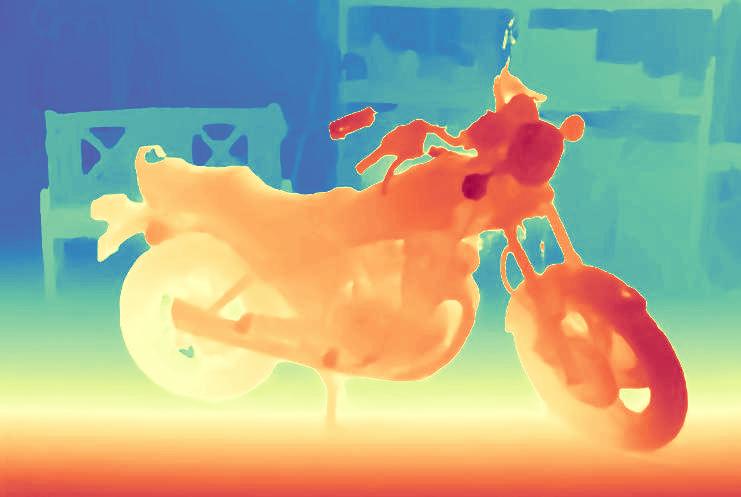}  &
\includegraphics[width=0.18\textwidth]{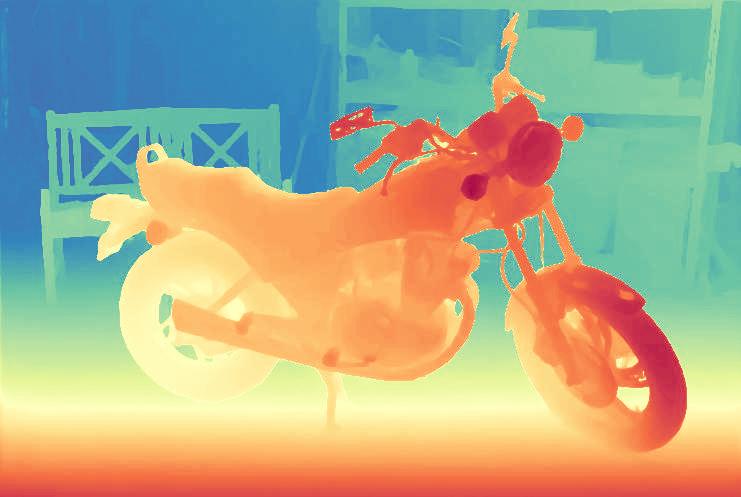}  \\
\includegraphics[width=0.18\textwidth]{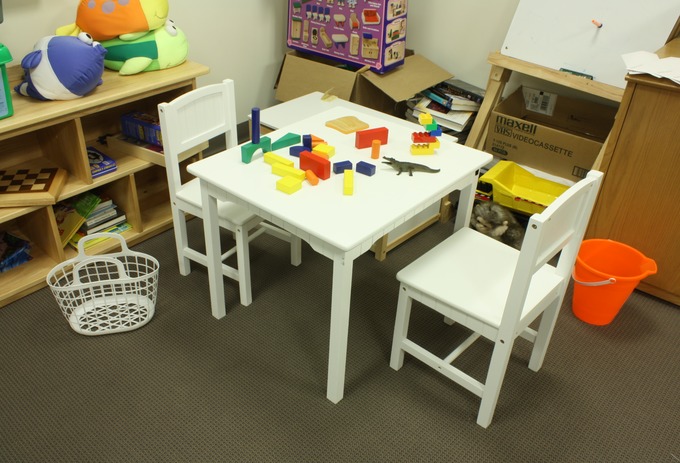} &
\includegraphics[width=0.18\textwidth]{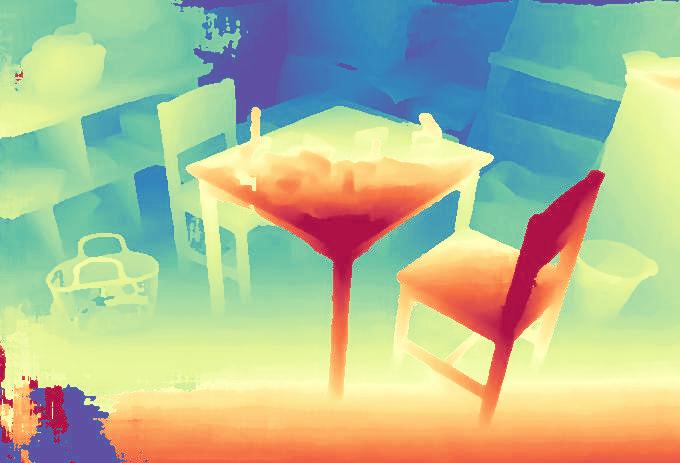} & 
\includegraphics[width=0.18\textwidth]{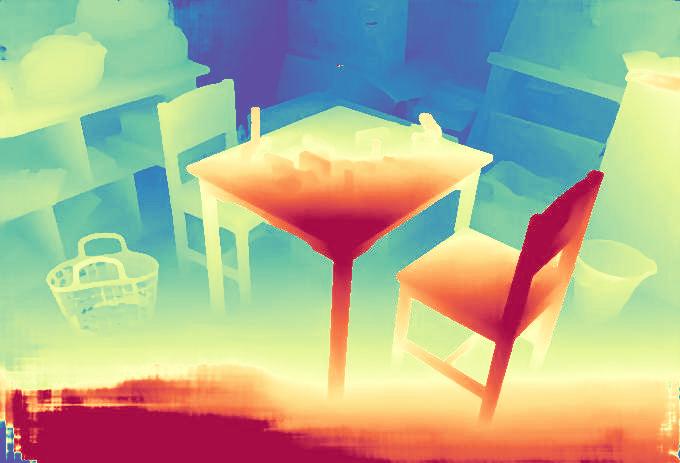} & 
\includegraphics[width=0.18\textwidth]{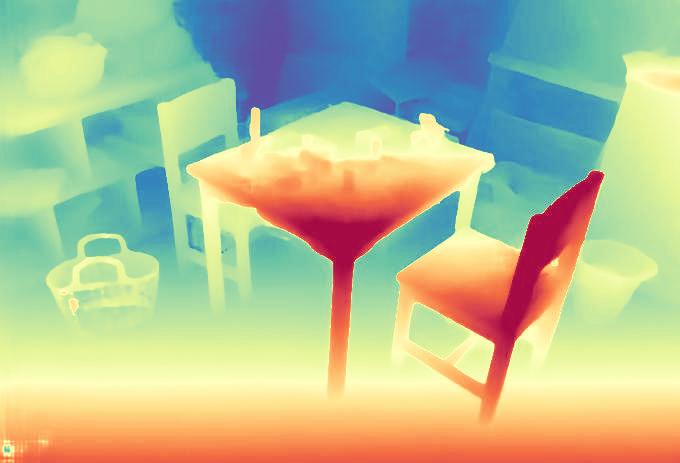}  &
\includegraphics[width=0.18\textwidth]{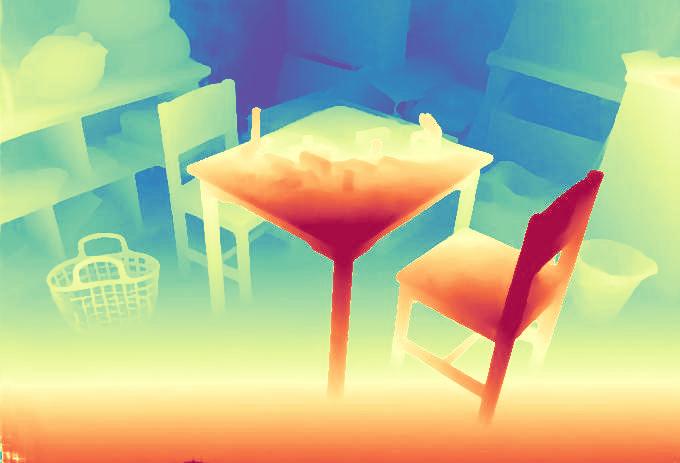}  \\
Reference image \cite{scharstein2014high} & Graft-PSMNet \cite{Liu_2022_CVPR_GraftNet} & ITSA-PSMNet \cite{Chuah_2022_CVPR_ITSA} & LSSI-PSMNet \cite{watson2020learning_LSSI} & NS-PSMNet \cite{Tosi_2023_CVPR} \\
\end{tabular}
\vspace{-0.3cm}
\caption{\textbf{Qualitative comparison -- PSMNet variants.} From left to right: reference image, disparity maps predicted by networks trained on synthetic data with ground-truth (Graft-PSMNet, ITSA-PSMNet) or on real data without any ground-truth (MfS-PSMNet, NS-PSMNet).} 
\label{fig:image_collection8}
\end{figure*}

Zero-shot generalization refers to the capability of a stereo network to generalize from one domain (\eg, synthetic data) to a completely different domain (\eg, real-world scenes) without the need for fine-tuning or adaptation. This is particularly desirable when collecting stereo images or ground truth data for the target domain is expensive or infeasible. 
It is worth mentioning that the recent iterative models reviewed in Sec. \ref{sec:iterative_optimization-based} excel at this \cite{Tosi_2023_CVPR}, despite not implementing any strategy specific for this purpose.


\paragraph{Domain-Agnostic Feature Modeling}

These methods focus on various aspects, such as feature regularization, feature consistency, shortcut avoidance, and representation learning, each tackling the problem from a unique perspective while sharing the common goal of learning robust, domain-invariant representations.
\hypertarget{DSMNet}{\textbf{DSMNet}} [\href{https://github.com/feihuzhang/DSMNet}{Code}] \cite{zhang2019domaininvariant_DSM} and \hypertarget{FCStereo}{\textbf{FCStereo}} [\href{https://github.com/jiaw-z/FCStereo}{Code}] \cite{Zhang_2022_CVPR_FCStereo} both aim to learn domain-invariant features, but they differ in their approach. DSMNet introduces a Domain Normalization (DN) layer to regularize the distribution of learned features across spatial and channel dimensions, reducing sensitivity to image-level style variations and local contrast differences between domains. 
In contrast, FCStereo focuses on explicitly encouraging feature consistency between matching pixels from left and right views through two loss functions: the stereo contrastive feature (SCF) loss and the stereo selective whitening (SSW) loss. While DSMNet emphasizes feature regularization, FCStereo prioritizes feature consistency across domains.
\hypertarget{GraftNet}{\textbf{GraftNet}} [\href{https://github.com/SpadeLiu/Graft-PSMNet}{Code}] \cite{Liu_2022_CVPR_GraftNet} takes a different path by leveraging broad-spectrum features from a model pre-trained on large-scale datasets. It grafts these features to the cost aggregation module of an existing stereo network and uses a shallow network to restore task-related
information.
In comparison to DSMNet and FCStereo, which learn domain-invariant features from scratch, GraftNet exploits existing knowledge from pre-trained models to improve generalization.
\hypertarget{ITSA}{\textbf{ITSA}} [\href{https://github.com/waychin-weiqin/ITSA}{Code}] \cite{Chuah_2022_CVPR_ITSA} and \hypertarget{HVT}{\textbf{HVT}} [\href{https://github.com/cty8998/HVT-PSMNet}{Code}] \cite{Chang_2023_CVPR} both address the issue of shortcut learning, where networks exploit spurious correlations or superficial cues in synthetic training data rather than learning transferable representations. On the one hand, ITSA uses information-theoretic losses to automatically restrict the encoding of shortcut-related information into feature representations by combining the task loss with an approximation of the Fisher information loss. 
On the other hand, HVT emphasizes data augmentation,  by transforming synthetic training images hierarchically at global, local, and pixel levels to diversify the training domain and prevent the model from learning dataset-dependent shortcuts. 
\hypertarget{MRL-Stereo}{\textbf{MRL-Stereo}} \cite{Rao_2023_CVPR}, instead, introduces an approach to learning domain-invariant representations using masked representation learning. By feeding a masked left image and a complete right image into the model and reconstructing the original left image, MRL-Stereo encourages the learning of structural and domain-invariant features. 


\paragraph{Non-parametric Cost Volumes}
Non-parametric cost volume construction methods build cost volumes using conventional, domain-agnostic matching functions rather than relying on learned features that may be sensitive to domain-specific characteristics.  Matching-Space Networks (\hypertarget{MS-Nets}{\textbf{MS-Nets}}) [\href{https://github.com/ccj5351/MS-Nets}{Code}] \cite{cai2020matchingspace}, for example, move the learning process from the color space to the Matching Space by replacing learning-based feature extraction with four conventional matching functions (NCC, ZSAD, CENSUS, and SOBEL) and their associated confidence scores. These functions and scores are combined to generate a 4D matching volume, which is then regularized using adapted versions of popular architectures like GCNet \cite{kendall2017end_GC-NET} and PSMNet \cite{chang2018pyramid_PSMNet}. Similarly, \hypertarget{ARStereo}{\textbf{ARStereo}} [\href{https://github.com/kelkelcheng/AdversariallyRobustStereo}{Code}] \cite{cheng2022revisiting_AdversariallyRobustStereo} leverages Census Transform binary patterns as the matching features for building cost volumes, together with a backbone applied to extract high-level semantic contextual features from the reference image alone, avoiding the vulnerabilities of matching at features level. 
Then, this hybrid cost volume is regularized through the remaining layers.


\paragraph{Integration of Additional Geometric Cues}

Incorporating complementary geometric information, such as sparse depth hints or refined disparities from traditional algorithms, can guide stereo networks toward more robust and generalizable predictions. Two notable methods in this direction are Neural Disparity Refinement (\hypertarget{NDR}{\textbf{NDR}}) [\href{https://cvlab-unibo.github.io/neural-disparity-refinement-web/}{Code}] \cite{aleotti2021neural,tosi2024neural} and Expansion of Visual Hints for Stereo (\hypertarget{EVHS}{\textbf{EVHS}}) \cite{Pilzer_2023_WACV_EVHS}. On the one hand, NDR combines traditional stereo algorithms with deep learning to obtain refined, high-resolution disparity maps with sharp edges. Trained solely on synthetic data, the network can refine and super-resolve disparity maps from any source, including classical stereo methods or deep stereo networks, generalizing well to real images in a zero-shot manner. 
On the other hand, EVHS exploits visual hints from visual-inertial odometry. The method expands sparse and unevenly distributed 3D cues using a 3D random geometric graph, connecting hints that are close in the 3D world to improve the learning and inference process, filtered by confidence. 
Expanded hints are integrated within DeepPruner \cite{duggal2019deeppruner}, guiding the differentiable patchmatch algorithm within a narrow disparity range. 


\paragraph{Real-World Monocular to Synthetic Stereo Data}

Generating diverse, realistic training data is crucial for improving generalization. To address this, approaches like rendering synthetic data or using real-world single/sparse images have been proposed. Methods like Learning Stereo from Single Images (LSSI)  (\hypertarget{LSSI}{\textbf{LSSI}}) [\href{https://github.com/nianticlabs/stereo-from-mono/}{Code}] \cite{watson2020learning_LSSI} and NeRF-Supervised Deep Stereo (NS-Stereo) (\hypertarget{NS-Stereo}{\textbf{NS-Stereo}}) [\href{https://nerfstereo.github.io/}{Code}] \cite{Tosi_2023_CVPR} create diverse stereo training data directly from easily acquired real-world monocular images. This exposes stereo networks to natural textures/appearances, promoting zero-shot generalization, while enabling training on numerous scenes leveraging intrinsic real-world visual properties.
Specifically, LSSI leverages pre-trained monocular depth networks (\eg, MiDaS \cite{Ranftl2022}) to predict depth maps for single images, which are then converted to disparity maps and used both to synthesize stereo pairs, as well as pseudo-labels for training a stereo network. 
Similarly, NS-Stereo generates stereo training data from sparse real-world image sequences captured with a single handheld camera. However, instead of using monocular depth estimation, NS-Stereo fits a NeRF model \cite{muller2022instant} to each scene and uses it to render arbitrary stereo pairs by synthesizing a reference view and a target view displaced by a virtual baseline.
Interestingly, NS-Stereo takes a step further by generating stereo triplets to handle occlusions in the photometric loss and exploiting the depth maps rendered by NeRF as proxy supervision. 


\paragraph{Knowledge Transfer}


Fine-tuning a pre-trained network often harms generalization itself. This happens, in particular, when the original training dataset is much larger than the few samples used for fine-tuning, however, few works focused on overcoming this issue. On this path, \textbf{DKT-Stereo}
[\href{https://github.com/jiaw-z/DKT-Stereo}{Code}] \cite{zhang2024robust_DKT} \hypertarget{DKT-Stereo}{
aims at preserving the original, ``Dark Knowledge" of a model during fine-tuning, with a frozen teacher network, an exponential moving average (EMA) teacher network, and a student network, all initialized with the same pre-trained weights. 
A further Filter and Ensemble (F\&E) module discards the region being inconsistent across the pseudo-labels by the frozen and the EMA teachers to avoid insufficient regularization, and ensembles the labels in the consistent region to prevent overfitting ground-truth details at the expense of generalization. 
}


\subsubsection{Offline Adaptation}
\label{sec:offline-adaptation}

When a set of stereo pairs is available from an unseen domain, the domain shift can be compensated for by carrying out offline domain adaptation. 
In the absence of ground-truth annotations, a common approach consists of using self-supervised learning techniques, such as photometric losses, to either train from scratch a network specialized for the target domain or to fine-tune a pre-trained one. 

\textbf{Flow2Stereo} [\href{https://github.com/ppliuboy/Flow2Stereo}{Code}] \cite{Liu_2020_CVPR_Flow2Stereo} \hypertarget{Flow2Stereo} {pursues the former strategy by jointly learning optical flow and stereo matching by exploiting the 3D geometry of stereoscopic videos in two stages. 
In the first stage, a teacher network predicts confident optical flow using photometric consistency and geometric constraints. In the second stage, the student network is refined using a self-supervised loss and proxy learning tasks, with the teacher network's confident predictions serving as pseudo-labels.}
%
%
On the same track, \textbf{Reversing-Stereo} [\href{https://github.com/FilippoAleotti/Reversing}{Code}] \cite{aleotti2020reversing} \hypertarget{Reversing-Stereo}{reverses the typical relationship between monocular and stereo depth estimation \cite{monodepth17}. At the core, a monocular completion network (MCN) leverages sparse disparity points computed by any traditional stereo algorithm and single-image cues to predict dense and accurate disparity maps, 
by aggregating multiple MCN predictions with a randomized subset of the input points. 
These high-quality proxy labels are then used to supervise the training of deep stereo networks.} Similarly, \textbf{Revealing-Stereo} \cite{Chen_2021_ICCV} \hypertarget{Revealing-Stereo}{proposes a framework to improve both stereo and monocular depth estimation by leveraging their reciprocal relations, introducing an occlusion-aware distillation strategy to train a monocular depth network with reliable predictions from a stereo network, and an occlusion-aware fusion module to combine their advantages.} \textbf{TiO-Depth} [\href{https://github.com/ZM-Zhou/TiO-Depth_pytorch}{Code}] \cite{zhou2023two} \hypertarget{TiO-Depth}{, instead, is a two-in-one self-supervised model handling monocular and binocular tasks via a Siamese architecture with monocular feature matching and multi-stage joint training.}
%
%
\textbf{PASMNet} [\href{https://github.com/The-Learning-And-Vision-Atelier-LAVA/PAM}{Code}] \cite{shen2023digging} \hypertarget{PASMNet}{ deploys a parallax-attention mechanism (PAM) 
integrating epipolar constraints with an attention mechanism to calculate feature similarities along the epipolar line. 
PASMnet, instead, employs this mechanism within a feature extractor, a cascaded parallax-attention module for coarse-to-fine matching cost regression, and a disparity refinement module, and is trained with 
a combination of photometric loss, smoothness loss, and PAM-specific losses that enforce left-right consistency, smoothness, and cycle consistency of the parallax-attention maps at multiple scales.
}

\textbf{MultiscopicVision} \cite{yuan2021stereo} \hypertarget{MultiscopicVision}{proposes a self-supervised framework for stereo matching utilizing multiple images captured at aligned camera positions, introducing losses to optimize the network without ground-truth.} \textbf{StereoGAN} [\href{https://github.com/ruiliu-ai/StereoGAN}{Code}] \cite{Liu_2020_CVPR_StereoGAN} \hypertarget{StereoGAN}{tackles the domain gap between synthetic and real data by jointly optimizing a domain translation network and a stereo matching network. The domain translation component utilizes a GAN-based adversarial loss to generate realistic images from synthetic data while maintaining stereo consistency and epipolar geometry through novel constraints. By leveraging knowledge of the target domain, StereoGAN effectively adapts synthetic data to more closely resemble the characteristics of the real-world domain.}


\begin{figure*}[t]
    \centering
	\begin{subfigure}{0.22\linewidth}
		\includegraphics[width=\linewidth]{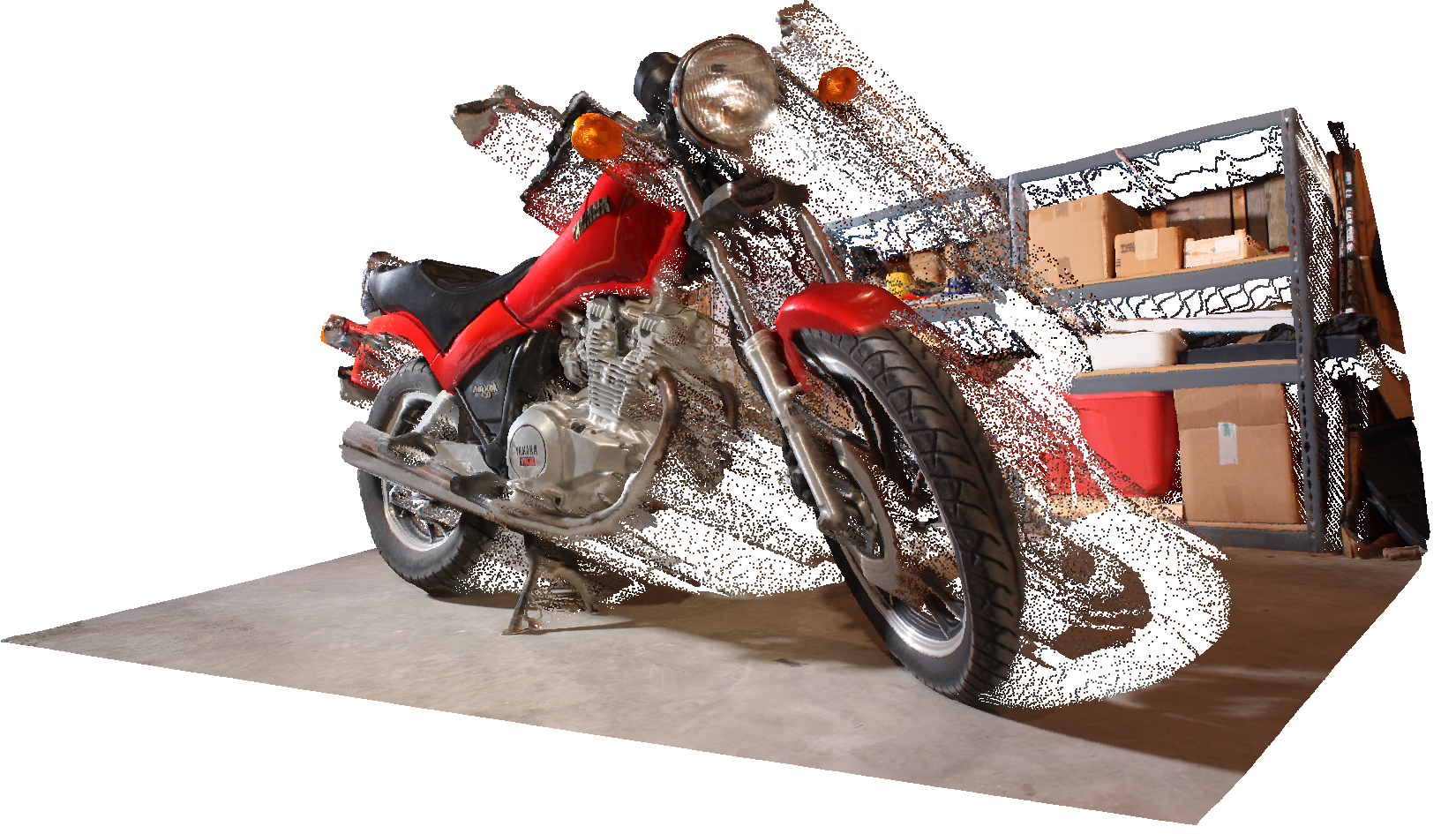}
		\caption{GM-Stereo}
		\label{fig:illustration_stereo_net}
	\end{subfigure}
	\begin{subfigure}{0.22\linewidth}
		\includegraphics[width=\linewidth]{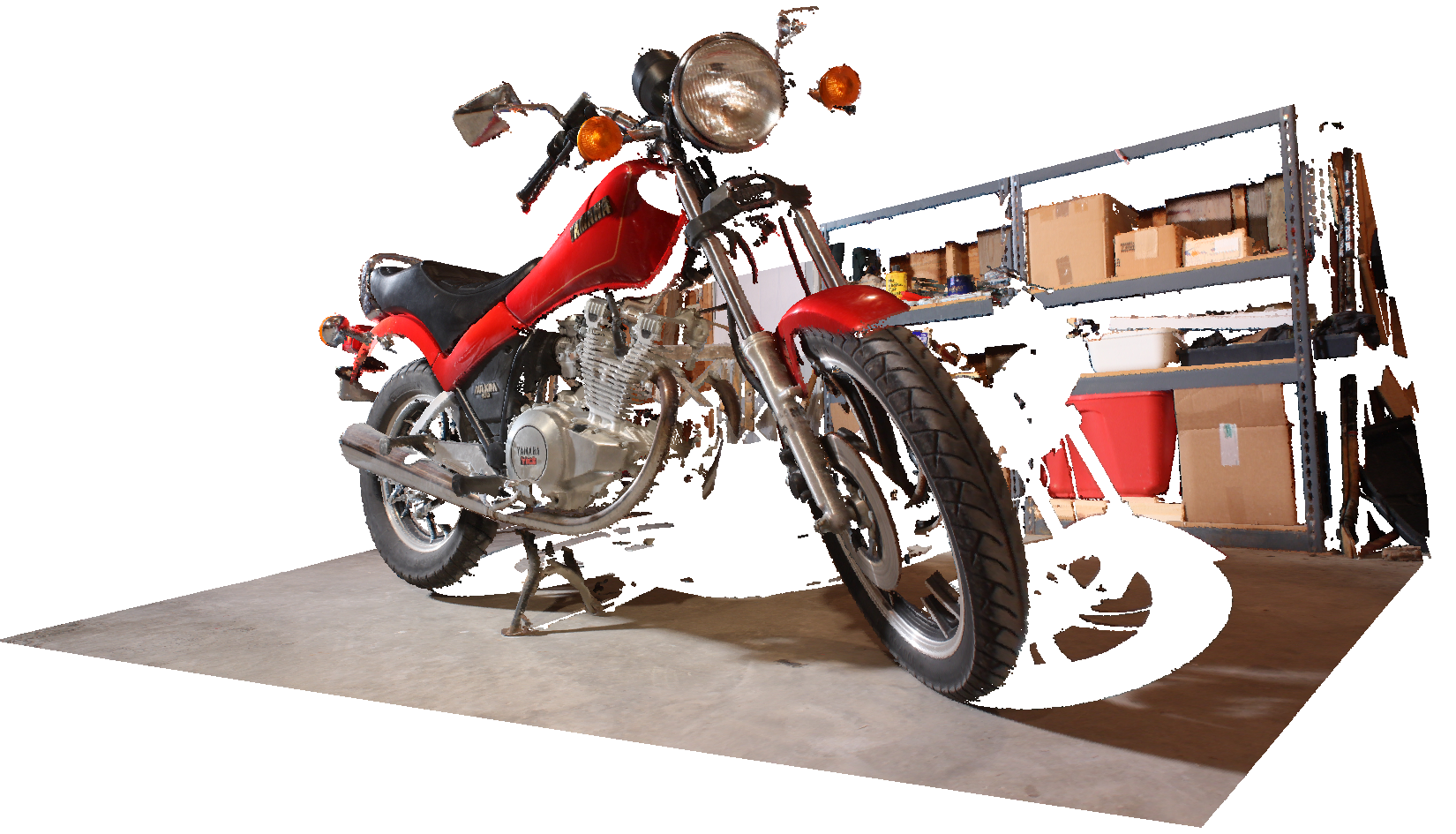}
		\caption{Ground Truth}
		\label{fig:illustration_stereo_net}
	\end{subfigure}
	\begin{subfigure}{0.22\linewidth}
		\includegraphics[width=\linewidth]{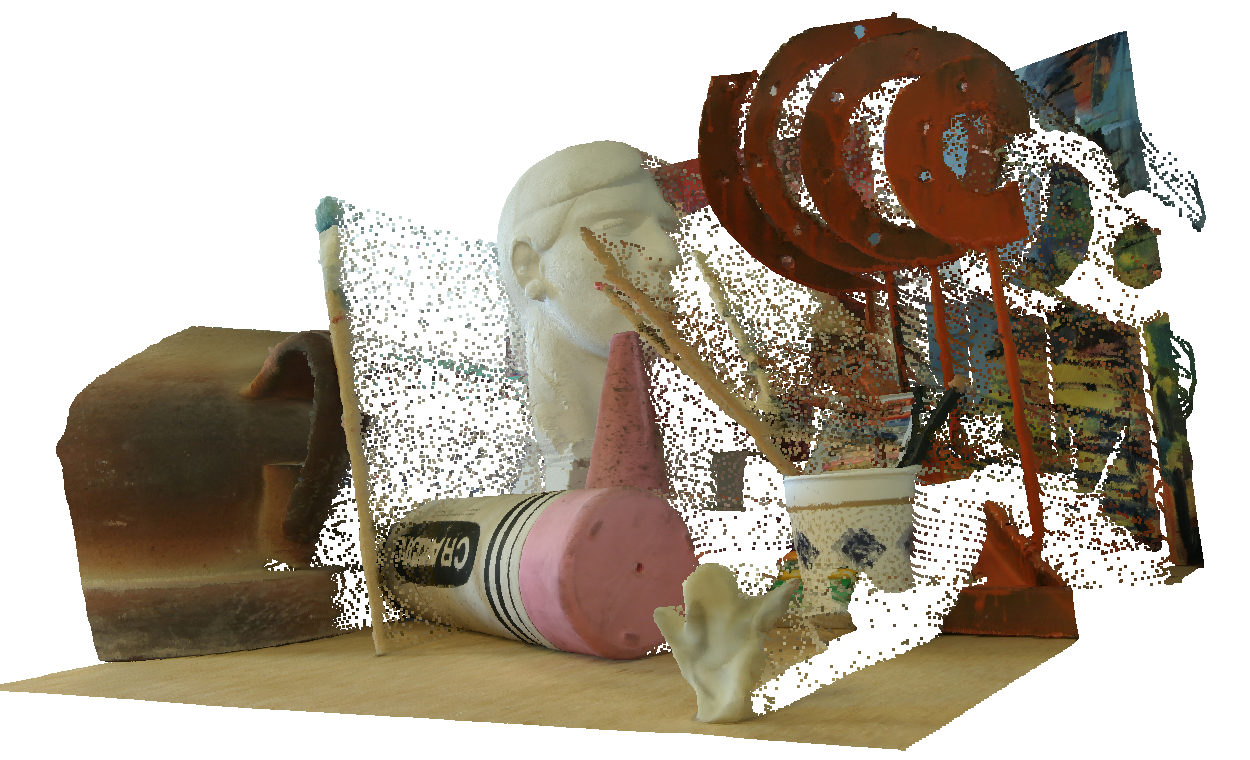}
        \caption{RAFT-Stereo}
		\label{fig:illustration_stereo_net}
	\end{subfigure}
	\begin{subfigure}{0.22\linewidth}
		\includegraphics[width=\linewidth]{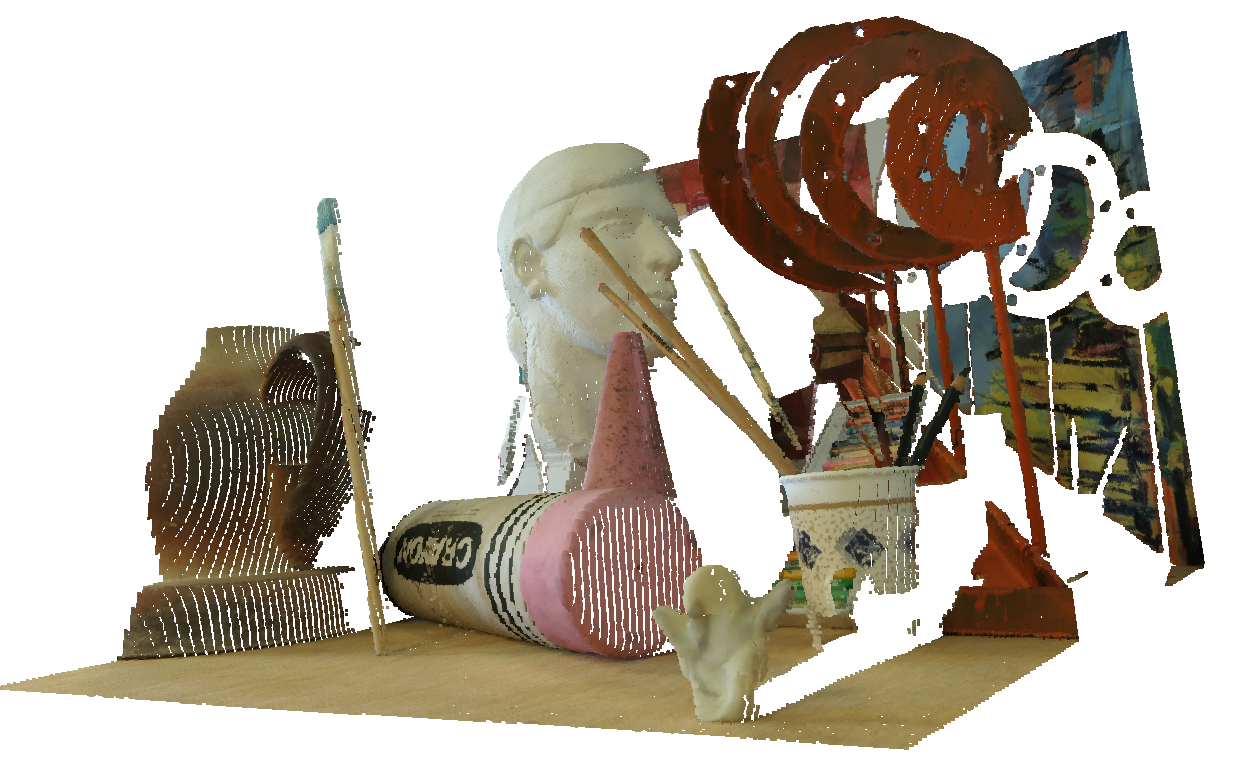}
		\caption{Ground Truth}
		\label{fig:illustration_stereo_net}
	\end{subfigure}
	\quad
	\vspace{-0.2cm}
	\caption{{\bf Bleeding Artifacts.} The smooth disparities predicted between foreground and background objects project into flying points in 3D space (a,c), whereas precise 3D reconstructions demand sharp discontinuities (b,d).  
	\vspace{-0.2cm}
	}
	\label{fig:illustration_bleeding}
\end{figure*}

Assuming a pre-trained model, \textbf{AdaStereo} \cite{Song_2021_CVPR_AdaStereo} \hypertarget{AdaStereo}{
aligns input representations and improves cross-domain adaptation ability, by deploying  
1) a non-adversarial progressive color transfer algorithm for input image-level alignment, 
2) an efficient, parameter-free cost normalization layer for channel-level feature alignment, 
to regulate the norm distribution of pixel-wise feature vectors; 
and 3) a self-supervised 
auxiliary task that reconstructs target-domain images using predicted disparities and occlusion masks. 
}


In contrast, \textbf{UCFNet} [\href{https://github.com/gallenszl/UCFNet?tab=readme-ov-file}{Code}] \cite{shen2023digging} \hypertarget{UCFNet}{
extends CFNet \cite{shen2021cfnet} and adapts a pre-trained model to the target domain using the pseudo-labels predicted by the model itself, and filtered according to uncertainty. 
}
Lastly, \textbf{RAG} [\href{https://github.com/chzhang18/RAG}{Code}] \cite{Zhang_2022_CVPR} \hypertarget{RAG}{exploits NAS for discovering novel cells and adapting a pre-trained network to a specific domain, by expanding its architecture. Accordingly, RAG discovers a different structure for each domain: at test time, the proper one is selected depending on the domain faced, by a Scene Router network ensembling a set of auto-encoders trained on the single domains during adaptation.
}


\subsubsection{Online Stereo Adaptation}
\label{sec:online-adaptation}
When stereo images from the target domain are not available beforehand, a stereo network can be adapted in an online manner during deployment \cite{Tonioni_2019_CVPR}. In this case, stereo image pairs from the target domain are collected continuously, and the network 
adapts by relying on self-supervised learning objectives, such as photometric consistency loss. 
The main challenges of this setting consist of avoiding degradation of the model and preserving efficiency. 

The first works extending \cite{Tonioni_2019_CVPR} are \textbf{AoHNet} \cite{wang2020faster} and \textbf{MAD++} [\href{https://github.com/CVLAB-Unibo/Real-time-self-adaptive-deep-stereo}{Code}] \cite{poggi2021continual}. Both enhance the effectiveness and speed of the adaptation procedure by replacing photometric losses with pseudo-labels from traditional stereo algorithms. In addition, AoHNet deploys an Adapt-or-Hold (AoH) mechanism to figure out whether to adapt or not, thus reducing the overall computational overhead.
%
%
%
%
%
%
%
%
In contrast, 
\textbf{PointFix} \cite{kim2022pointfix} \hypertarget{PointFix}{exploits Model Agnostic Meta-Learning (MAML) to improve online adaptation. In the inner loop, wrong pixels are selected from the prediction by the base network and fixed by a PointFixNet, which predicts residuals used to adapt the base network itself. In the outer loop, both the base network and the PointFixNet are optimized based on the performance of the former after adaptation.
At test time, only the inner loop is performed.}
%
%
%
Lastly, \textbf{FedStereo} [\href{https://github.com/mattpoggi/fedstereo}{Code}] \cite{Poggi_2024_CVPR_FedStereo} \hypertarget{FedStereo}{casts online adaptation as a distributed process, in which a server receives and aggregates updated weights from a set of actively adapting clients. These weights are then sent to a listening client not carrying out adaptation. To minimize communication overhead, a strategy inspired by MAD \cite{Tonioni_2019_CVPR} is used to transfer only a subset of the weights.}



\subsection{Over-Smoothing}
\label{sec:over-smoothing}

A common limitation of most stereo networks is their tendency to over-smooth depth discontinuities. 
During inference, 3D stereo architectures typically employ a $\textit{soft argmax}$ operation to obtain the final disparity estimate by calculating the mean of the often multimodal predicted distribution. 
This leads to disparity estimates falling between the foreground and background modes, resulting in erroneous predictions and over-smoothed depth discontinuities.  This approach leads to disparity estimates that fall between the foreground and background modes, resulting in erroneous predictions and over-smoothed depth discontinuities. 2D stereo architectures, despite not explicitly using soft argmax, are still affected by this issue. 
This yields inconsistent ``bleeding" artifacts in the reconstructed 3D geometry around object boundaries, 
highly undesirable for applications requiring accurate reconstruction with precise contours. 

\subsubsection{Unimodal Distribution Modeling}
\label{sec:unimodal}

Unimodal distribution modeling aims to alleviate the over-smoothing issue by constraining the disparity estimation to a single dominant mode. \hypertarget{SM-CDE}{\textbf{SM-CDE}} \cite{chen2019over_smoothing} introduces a single-modal weighted average operation during inference. It considers the locality of the estimated disparity distribution and applies weighted averaging only to the dominant mode. Additionally, SM-CDE analyzes different loss functions and demonstrates that using cross-entropy loss with Gaussian distribution during training provides more stable and fine-grained supervision. Similarly, \hypertarget{AcfNet}{\textbf{AcfNet}} [\href{https://github.com/youmi-zym/AcfNet}{Code}] \cite{zhang2020adaptive_AcfNet} aims to improve the sharpness of disparity maps by directly supervising the cost volume with adaptive unimodal ground truth distributions. It introduces a confidence estimation network to modulate the variance of the unimodal distribution based on the confidence of finding a unique match. This ensures that pixels with high confidence have sharp peaks, while those with low confidence have flatter peaks. AcfNet also proposes a stereo focal loss to address the sample imbalance problem in the cost volume. Another approach, \hypertarget{CDN}{\textbf{CDN}} [\href{https://github.com/Div99/W-Stereo-Disp}{Code}] \cite{garg2020wasserstein}, addresses the issue by introducing an architecture that outputs a continuous distribution over arbitrary disparity values. It predicts probabilities and real-valued offsets for each disparity value in a pre-defined discrete set and takes the mode of this distribution as the final prediction. 
CDN also uses a novel loss function based on the Wasserstein distance between the true and predicted distributions.
While the above methods focus on modeling unimodal distributions, \hypertarget{LaC}{\textbf{LaC}} [\href{https://github.com/SpadeLiu/Lac-GwcNet}{Code}] \cite{liu2022local_LaC} 
introduces a Local Similarity Pattern (LSP) to capture local structural information by explicitly revealing relationships between a point and its neighbors. To address the over-smoothing problem caused by static convolutional filters, LaC proposes Cost Self-Reassembling (CSR) and Disparity Self-Reassembling (DSR) strategies to adaptively propagate reliable disparity values based on image content. 

\subsubsection{Multi-Modal Distribution Modeling}
\label{sec:multimodal}

Moving beyond unimodal distributions, multi-modal distribution modeling has emerged as a promising approach to achieve sharp depth discontinuities. \hypertarget{SMD-Nets}{\textbf{SMD-Nets}} [\href{https://github.com/fabiotosi92/SMD-Nets}{Code}] \cite{Tosi2021CVPR_SMD} is a pioneering work in this direction, utilizing bimodal mixture densities as output representation. It encodes the input into a feature map, from which a multi-layer perceptron estimates the parameters of a bimodal Laplacian mixture distribution at any continuous 2D location.  The bimodal distribution effectively models both the foreground and background disparities, allowing for precise depth estimation near object boundaries. The final disparity is obtained by selecting the mode with the highest density value. Building upon this concept, \hypertarget{ADL}{\textbf{ADL}} [\href{https://github.com/xxxupeng/ADL}{Code}] \cite{Xu_CVPR_2024_ADL} proposes an adaptive multi-modal cross-entropy loss that models the ground-truth disparity distribution as a mixture of Laplacians. The number of modes and their weights are determined by local clustering and statistics within a local window. ADL also introduces a dominant-modal disparity estimator (DME) to robustly locate the dominant mode. 


\subsection{Transparent and Reflective Objects}
\label{sec:tom}

Non-Lambertian materials have always posed significant challenges to matching algorithms, as they introduce misleading visual information about scene geometry. 
When a transparent object is present, a matching algorithm computes correspondences between points behind it, failing to perceive the object's distance from the camera. Alternatively, when dealing with a reflective surface, the algorithm might fail to triangulate depth properly. 
However, driven by data, deep learning has the potential to address these difficult challenges as well.
In this regard, \textbf{DDF} \cite{chai2020deep} \hypertarget{DDF}{proposes a framework for fusing the disparity predicted from stereo images with the depth acquired by a structured-light camera. The fusion is performed as a per-pixel weighted average of the two, with the weights predicted by a 2D UNet. An additional UNet is deployed for refining the fused depth map.}
An alternative strategy consists of segmenting the non-Lambertian materials and use the semantic masks to assist the stereo network. 
\textbf{TA-Stereo} \cite{wu2023transparent} and
\textbf{Depth4ToM} [\href{https://github.com/CVLAB-Unibo/Depth4ToM-code}{Code}] \cite{costanzino2023learning} implement this approach from two different perspectives. \hypertarget{TA-Stereo}{The former applies the segmentation masks to the stereo images to enforce similar appearance across the two and directly ease matching at testing time.} \hypertarget{Depth4ToM}{In contrast, Depth4Tom 
in-paints non-Lambertian objects according to the segmentation masks and processes them to obtain pseudo-labels for fine-tuning the stereo model. 
Pseudo-labels are predicted by a pre-trained monocular depth model and are fused with the predictions by the stereo network itself, replacing these latter in correspondence with non-Lambertian materials. 
}


\subsection{Asymmetric Stereo}
\label{sec:asymmetric}
Most stereo frameworks assume the stereo image pair is captured by cameras with identical properties. However, asymmetries between the images are common, due to differences in the cameras' intrinsic parameters, resolutions, or noise levels. These asymmetries can affect the matching process between the two images. Consequently, a new research direction has emerged to design stereo solutions that are robust to such asymmetries between the stereo pairs.

Visually-Imbalanced Stereo (\textbf{VI-Stereo})\cite{liu2020visually} \hypertarget{Visually-Imbalanced Stereo}{is the first work in this direction. Given an HR left and LR right image pair, it deploys a UNet to upsample the latter and restore a balanced stereo pair, which is processed by a DispNet \cite{mayer2016large} to predict an HR disparity map.}
%
%
On the contrary, \textbf{NDR} [\href{https://github.com/CVLAB-Unibo/neural-disparity-refinement}{Code}] \cite{aleotti2021neural,tosi2024neural} \hypertarget{NDR}{downsamples the left image to the resolution of the right one, runs a traditional stereo algorithm to obtain an LR and then upsample it, guided by the HR features, according to a continuous formulation. }
%
%
Differently from the previous works, \textbf{DA-AS} \cite{Chen_2022_CVPR} \hypertarget{DA-AS}{studies a self-supervised setting and proposes feature-metric consistency to replace the photometric loss, which yields sub-optimal results with asymmetric images, yet can be used an initial stereo model to learn good features for matching. Then, these features can be used to compute feature-metric consistency loss and train a new model starting from it.}
%
%
Lastly, \textbf{SASS} \cite{song2023unsupervised} \hypertarget{SASS}{proposes a novel spatially-adaptive self-similarity measure in a self-supervised manner, by extending the concept of self-similarity to generate deep features that are robust to the asymmetries by leveraging contrastive learning.}


\begin{table}[t]
\centering
\setlength{\tabcolsep}{10pt}
\rowcolors{2}{salmon}{white}
\scalebox{0.65}{
\begin{tabular}{ll cccc}

\Xhline{2pt}
\multirow{2}{*}{Method}  & \multirow{2}{*}{Venue} & \multicolumn{4}{c}{KITTI 2015}\\
\cmidrule(lr){3-6}
& &  D1-bg $\downarrow$  & D1-fg $\downarrow$ & D1-all $\downarrow$ & Time (s)\\


\Xhline{2pt}
\multicolumn{6}{c}{\hyperlink{sec:architecture}{Deep Stereo Networks (2020-2024)}} \\
\hline

\hyperlink{MoCha-Stereo}{\textcolor{magenta}{MoCha-Stereo}}~\cite{mocha}                               & CVPR 2024    & \bronze{1.36} & \bronze{2.43} & \gold{\textcolor{purple}{1.53}} & 0.27\\
\hyperlink{ADL}{\textcolor{magenta}{GANet+ADL}}~\cite{Xu_CVPR_2024_ADL}                            & CVPR 2024    & 1.38 & \silver{2.38} & \silver{1.55} & 0.67\\
\hyperlink{Selective-Stereo}{\textcolor{magenta}{Selective-IGEV}}~\cite{wang2024selective}         & CVPR 2024    & \silver{1.33} & 2.61 & \silver{1.55} & 0.24\\
\hyperlink{MC-Stereo}{\textcolor{magenta}{MC-Stereo}}~\cite{Feng_2024_3DV_MC-Stereo}               & 3DV 2024     & \bronze{1.36} & 2.51 & \silver{1.55} & 0.40\\
\hyperlink{Any-Stereo}{\textcolor{magenta}{Any-IGEV}}~\cite{Liang_Li_2024_Any-Stereo}              & AAAI 2024    & 1.43 & \gold{\textcolor{purple}{2.35}} & 1.58 & 0.32\\
\hyperlink{NMRF}{\textcolor{magenta}{NMRF-Stereo}}~\cite{guan2024neural_NMRF}                      & CVPR 2024    & \gold{\textcolor{purple}{1.28}} & 3.13 & 1.59 & 0.09\\
\hyperlink{CroCo-Stereo}{\textcolor{magenta}{CroCo\_RVC}}~\cite{croco_v2}                          & ICCV 2023    & 1.38 & 2.65 & 1.59 & 0.93\\
\hyperlink{IGEV-Stereo}{\textcolor{magenta}{IGEV-Stereo}}~\cite{xu2023iterative_IGEV-Stereo}       & CVPR 2023    & 1.38 & 2.67 & 1.59 & 0.18\\
\hyperlink{LEAStereo}{\textcolor{magenta}{LEAStereo}}~\cite{cheng2020hierarchical_LEAStereo}       & NeurIPS 2020 & 1.40 & 2.91 & 1.65 & 0.30\\
\hyperlink{ACVNet}{\textcolor{magenta}{ACVNet}}~\cite{Xu_2022_CVPR_ACVNet}                         & CVPR 2022    & 1.37 & 3.07 & 1.65 & 0.2\\
\hyperlink{PCW-Net}{\textcolor{magenta}{PCW-Net}}~\cite{shen2022pcw}                               & ECCV 2022    & 1.37 & 3.16 & 1.67 & 0.44\\
\hyperlink{LaC}{\textcolor{magenta}{LaC+GANet}}~\cite{liu2022local_LaC}                            & AAAI 2022    & 1.44 & 2.83 & 1.67  & 1.8\\
\hyperlink{CREStereo}{\textcolor{magenta}{CREStereo}}~\cite{li2022practical_CREStereo}             & CVPR 2022    & 1.45 & 2.86 & 1.69 & 0.41\\
\hyperlink{Any-Stereo}{\textcolor{magenta}{Any-RAFT}}~\cite{Liang_Li_2024_Any-Stereo}              & AAAI 2024    & 1.44 & 3.04 & 1.70 & 0.34\\
\hyperlink{DKT-Stereo}{\textcolor{magenta}{DKT-IGEV}}~\cite{zhang2024robust_DKT}                   & CVPR 2024    & 1.46 & 3.05 & 1.72 & 0.18\\
\hyperlink{GMStereo}{\textcolor{magenta}{GMStereo}}~\cite{xu2023unifying_GMStereo}               & TPAMI 2023   & 1.49 & 3.14 & 1.77 & 0.17\\

\hyperlink{GOAT}{\textcolor{magenta}{GOAT}}~\cite{liu2024global_GOAT}                              & WACV 2024    & 1.71 & 2.51 & 1.84 & 0.29\\
\hyperlink{CFNet}{\textcolor{magenta}{CFNet}}~\cite{shen2021cfnet}                                 & CVPR 2021    & 1.54 & 3.56 & 1.88 & 0.18\\
\hyperlink{AcfNet}{\textcolor{magenta}{AcfNet}}~\cite{zhang2020adaptive_AcfNet}                    & AAAI 2020    & 1.51 & 3.80 & 1.89 & 0.48\\
\hyperlink{CDN}{\textcolor{magenta}{CDN}}~\cite{garg2020wasserstein}                               & NeurIPS 2020 & 1.66 & 3.20 & 1.92 & 0.4\\
\hyperlink{AANet}{\textcolor{magenta}{AANet+}}~\cite{xu2020aanet}                                  & CVPR 2020    & 1.65 & 3.96 & 2.03 & 0.06\\
\hyperlink{SMD-Nets}{\textcolor{magenta}{PSM + SMD-Nets}}~\cite{Tosi2021CVPR_SMD}                  & CVPR 2021    & 1.69 & 4.01 & 2.08 & 0.41\\
\hyperlink{MSNet3D}{\textcolor{magenta}{MSNet3D}}~\cite{Shamsafar_2022_WACV_MobileStereoNet}       & WACV 2022    & 1.75 & 3.87 & 2.10 & 1.5\\
\hyperlink{Bi3D}{\textcolor{magenta}{Bi3D}}~\cite{Badki_2020_CVPR_Bi3D}                            & CVPR 2020    & 1.95 & 3.48 & 2.21 & 0.48\\
\hyperlink{AANet}{\textcolor{magenta}{AANet}}~\cite{xu2020aanet}                                   & CVPR 2020    & 1.99 & 5.39 & 2.55 & 0.062\\
\hyperlink{WaveletStereo}{\textcolor{magenta}{WaveletStereo}}~\cite{Yang_2020_CVPR_WaveletStereo}  & CVPR 2020    & 2.24 & 4.62 & 2.63 & 0.27\\

\hyperlink{AdaStereo}{\textcolor{magenta}{AdaStereo}}~\cite{Song_2021_CVPR_AdaStereo}              & CVPR 2021    & 2.59 & 5.55 & 3.08 & 0.41\\
\hyperlink{DSMNet}{\textcolor{magenta}{DSMNet-synthetic}}~\cite{zhang2019domaininvariant_DSM}      & ECCV 2020    & 3.11 & 6.72 & 3.71 & 1.6\\

\Xhline{2pt}
\multicolumn{6}{c}{\hyperlink{sec:efficiency}{Efficiency-Oriented Deep Stereo Networks (2020-2024)}} \\
\hline

\hyperlink{TemporalStereo}{\textcolor{magenta}{TemporalStereo}}~\cite{Zhang2023TemporalStereo}     & IROS 2023    & \silver{1.61} & \gold{2.78} & \gold{1.81} & 0.04\\
\hyperlink{PCVNet}{\textcolor{magenta}{PCVNet}}~\cite{zeng2023parameterized_PCVNet}                & ICCV 2023    & \bronze{1.68} & \silver{3.19} & \bronze{1.93} & {0.05}\\
\hyperlink{HITNet}{\textcolor{magenta}{HITNet}}~ \cite{Tankovich_2021_CVPR_HITNet}                 & CVPR 2021    & 1.74 & \bronze{3.20} & \bronze{1.98} & 0.02\\
\hyperlink{CasStereo}{\textcolor{magenta}{CasStereo}}~\cite{Gu_2020_CVPR_CasStereo}                & CVPR 2020    & \gold{1.59} & 4.03 & 2.00 & 0.6 \\
\hyperlink{CoEX}{\textcolor{magenta}{CoEX}}~\cite{bangunharcana2021correlate_CoEX}                 & IROS 2021    & 1.74 & 3.41 & 2.02 & 0.027\\
\hyperlink{BGNet}{\textcolor{magenta}{BGNet+}}~\cite{xu2021bilateral_BGNet}                        & CVPR 2021    & 1.81 & 4.09 & 2.19 & 0.03\\
\hyperlink{MABNet}{\textcolor{magenta}{MABNet\_origin}}~\cite{xing2020mabnet}                      & ECCV 2020    & 1.89 & 5.02 & 2.41 & 0.38\\
\hyperlink{Fast DS-CS}{\textcolor{magenta}{Fast DS-CS}}~\cite{Yee_2020_WACV_FDCSC}                 & WACV 2020    & 2.83 & 4.31 & 3.08 & 0.02\\
\hyperlink{MABNet}{\textcolor{magenta}{MABNet\_tiny}}~\cite{xing2020mabnet}                        & ECCV 2020    & 3.04 & 8.07 & 3.88 & 0.11\\
\hyperlink{StereoVAE}{\textcolor{magenta}{StereoVAE}}~\cite{chang2023stereovae}                    & ICRA 2023    & 4.25 & 10.18 & 5.23 & 0.03\\
\hyperlink{AAFS}{\textcolor{magenta}{AAFS}}~\cite{Chang_2020_ACCV_AAFS}                            & ACCV 2020    & 6.27 & 13.95 & 7.54 & 0.01\\

\Xhline{2pt}
\multicolumn{6}{c}{\href{https://ieeexplore.ieee.org/document/9395220}{Deep Stereo Networks (2016-2019)}} \\
\hline

CSPN \cite{cheng2019learning} & TPAMI 2019    & \silver{1.51}  & \gold{2.88} & \gold{1.74}  & 1.0\\
GANet \cite{zhang2019ga} & CVPR 2019    & \gold{1.48}  & \silver{3.46} & \silver{1.81}  & 1.8\\
GwcNet \cite{guo2019group_GWCNet} & CVPR 2019    & \bronze{1.74}  & 3.93 & \bronze{2.11}  & 0.32\\
HSMNet \cite{yang2019hierarchical_HSM} & CVPR 2019    & 1.80  & \bronze{3.85} & 2.14  & 0.14\\
PSMNet \cite{chang2018pyramid_PSMNet} & CVPR 2018    & 1.86  & 4.62 & 2.32  & 0.41\\
GC-Net \cite{kendall2017end_GC-NET} & ICCV 2017    & 2.21  & 6.16 & 2.87   & 0.9\\
DispNetC \cite{mayer2016large} & CVPR 2016    & 4.32  & 4.41 & 4.34   & 0.06\\

\bottomrule

\end{tabular}
}
\caption{\textbf{KITTI 2015 Online Benchmark.}}
\label{table:kitti_15}
\end{table}

\section{Experiments and Analysis}
\label{sec:experiments}

In this section, we compare the stereo frameworks surveyed so far on the standard benchmarks in the field.  Specifically, we report the leaderboards of the KITTI 2015 and the Middlebury v3 online benchmarks -- the former being the most popular dataset used in the literature, while the latter still represented a challenging benchmark in the late 2010s.
Additionally, we include results from the Robust Vision Challenge (ROB), which demonstrates the recent efforts to achieve robustness to domain shifts. Finally, we present the leaderboard of the Booster online benchmark, representing the challenges that remain open in the 2020s.
In each table, we highlight the \gold{first}, \silver{second}, and \bronze{third} best results in single categories, with the absolute best being reported in \textcolor{purple}{\textbf{purple}}.

\begin{table}[t]
\centering
\setlength{\tabcolsep}{8pt}
\rowcolors{2}{salmon}{white}
\scalebox{0.65}{
\begin{tabular}{llr r rr rr}

\Xhline{2pt}
\multirow{2}{*}{Method}  & \multirow{2}{*}{Venue} & \multirow{2}{*}{Res} & \multicolumn{2}{c}{Non-Occ} & \multicolumn{2}{c}{All} \\
\cmidrule(lr){4-5} \cmidrule(lr){6-7}
& & &  \textbf{bad2.0 $\downarrow$}  & EPE $\downarrow$ & bad2.0 $\downarrow$ & EPE $\downarrow$ \\

\Xhline{2pt}
\multicolumn{7}{c}{\hyperlink{sec:architectures}{Deep Stereo Networks (2020-2024)}} \\
\hline

\hyperlink{Selective-IGEV}{\textcolor{magenta}{Selective-IGEV}}~\cite{wang2024selective}                            & CVPR 2024    & F & \gold{\textcolor{purple}{2.51}} & \gold{\textcolor{purple}{0.91}} & \gold{\textcolor{purple}{6.04}} & \gold{\textcolor{purple}{1.54}} \\
\hyperlink{DLNR}{\textcolor{magenta}{DLNR}}~\cite{zhao2023high_DLNR}                               & CVPR 2023    & F & \silver{3.20} & \silver{1.06} & \silver{6.98} & \silver{1.91} \\
\hyperlink{EAI-Stereo}{\textcolor{magenta}{EAI-Stereo}}~\cite{zhao2022eai}                         & ACCV 2022    & F & \bronze{3.68} & \bronze{1.09} & \bronze{7.53} & \bronze{1.92}\\
\hyperlink{CREStereo}{\textcolor{magenta}{CREStereo}}~\cite{li2022practical_CREStereo}             & CVPR 2022    & F & 3.71 & 1.15 & 8.13 & 2.10\\
\hyperlink{RAFT-Stereo}{\textcolor{magenta}{RAFT-Stereo}}~\cite{lipson2021raft}                    & 3DV 2021     & F & 4.74 & 1.27 & 9.37 & 2.71\\
\hyperlink{IGEV-Stereo}{\textcolor{magenta}{IGEV-Stereo}}~\cite{xu2023iterative_IGEV-Stereo}       & CVPR 2023    & F & 4.83 & 2.89 & 8.16 & 3.64\\
\hyperlink{HIT-Net}{\textcolor{magenta}{HIT-Net}}~ \cite{Tankovich_2021_CVPR_HITNet}               & CVPR 2021    & H & 6.46 & 1.71 & 12.8 & 3.29\\
\hyperlink{GMStereo}{\textcolor{magenta}{GMStereo}}~\cite{xu2023unifying_GMStereo}                 & TPAMI 2023   & H & 7.14 & 1.31 & 11.7 & 1.89\\
\hyperlink{LEAStereo}{\textcolor{magenta}{LEAStereo}}~\cite{cheng2020hierarchical_LEAStereo}       & NeurIPS 2020 & H & 7.15 & 1.43 & 12.1 & 2.89\\
\hyperlink{CroCo-Stereo}{\textcolor{magenta}{CroCo-Stereo}}~\cite{croco_v2}                        & ICCV 2023    & F & 7.29 & 1.76 & 11.1 & 2.36\\
\hyperlink{PCVNet}{\textcolor{magenta}{PCVNet}}~\cite{zeng2023parameterized_PCVNet}                & ICCV 2023    & H & 8.19 & 1.53 & 13.6 & 2.71\\
\hyperlink{GOAT}{\textcolor{magenta}{GOAT}}~\cite{liu2024global_GOAT}                              & WACV 2024    & H & 8.73 & 1.64 & 13.8 & 2.71\\
\hyperlink{ACVNet}{\textcolor{magenta}{ACVNet}}~\cite{Xu_2022_CVPR_ACVNet}                         & CVPR 2022    & H & 13.6 & 8.24 & 19.5 & 12.1\\
\hyperlink{AdaStereo}{\textcolor{magenta}{AdaStereo}}~\cite{Song_2021_CVPR_AdaStereo}              & CVPR 2021    & F & 13.7 & 2.22 & 19.8 & 3.39 \\
\hyperlink{AANet}{\textcolor{magenta}{AANet++}}~\cite{xu2020aanet}                                 & CVPR 2020    & H & 15.4 & 6.37 & 22.0 & 9.77\\
\hyperlink{CasStereo}{\textcolor{magenta}{CasStereo}}~\cite{Gu_2020_CVPR_CasStereo}                & CVPR 2020    & H & 18.8 & 4.50 & 26.0 & 8.98\\

\Xhline{2pt}
\multicolumn{7}{c}{\href{https://ieeexplore.ieee.org/document/9395220}{Deep Stereo Networks (2016-2019)}} \\
\hline

HSM-Smooth-Occ \cite{yang2019hierarchical_HSM}                                                     & CVPR 2019    & F & \silver{10.8} & \gold{2.15} & \gold{17.5} & \gold{3.44}\\
MC-CNN-acrt \cite{vzbontar2016stereo}                                                              & JMLR 2016    & H & \gold{8.08} & \bronze{3.82} & \silver{19.1} & 17.9 \\
GANetREF\_RVC\cite{zhang2019ga}                                                                    & CVPR 2019    & H & 18.9 & 12.2 & \bronze{24.9} & 15.8 \\
SGM~\cite{hirschmuller2007stereo}                                                                  & TPAMI 2007   & H & \bronze{18.4} & 5.32 & 25.7 & 9.27\\
iResNet \cite{liang2018learning_iResNet}                                                           & CVPR 2018    & H & 22.9 & \silver{3.31} & 29.5 & \silver{4.67}\\
DeepPruner\_ROB \cite{duggal2019deeppruner}                                                        & ICCV 2019    & Q & 30.1 & 4.80 & 36.4 & \bronze{6.56}\\
PSMNet\_ROB \cite{chang2018pyramid_PSMNet}                                                         & CVPR 2018    & Q & 42.1 & 6.68 & 47.2 & 8.78\\

\bottomrule

\end{tabular}
}
\caption{\textbf{Middlebury-v3 Online Benchmark.}}
\label{table:middlebury_v3}
\end{table}

\begin{table*}[t]
\centering
\setlength{\tabcolsep}{10pt}
\rowcolors{2}{salmon}{white}
\scalebox{0.65}{
\begin{tabular}{ll cccc cccc cccc c}

\Xhline{2pt}
\multirow{2}{*}{Method}  & \multirow{2}{*}{Venue} & \multicolumn{4}{c}{KITTI 2015} & \multicolumn{4}{c}{Middlebury} & \multicolumn{4}{c}{ETH3D} & \multirow{2}{*}{Overall Rank}\\
\cmidrule(lr){3-6} \cmidrule(lr){7-10} \cmidrule(lr){11-14}
& &  D1-bg $\downarrow$  & D1-fg $\downarrow$ & D1-all $\downarrow$ & Rank  & bad1$\downarrow$ & bad2 $\downarrow$ & EPE $\downarrow$ & Rank & bad1$\downarrow$ & bad2 $\downarrow$ & EPE $\downarrow$ & Rank & \\

\Xhline{2pt}
\multicolumn{15}{c}{\href{http://robustvision.net/}{Robust Vision Challenge 2022}} \\
\hline

\hyperlink{CREStereo++}{\textcolor{magenta}{CREStereo++\_RVC}}~\cite{Jing_2023_ICCV_CREStereo++}  & ICCV 2023     & \gold{1.55} & \bronze{3.53} & \silver{1.88} & 2        &  \gold{\textcolor{purple}{16.5}} & \gold{\textcolor{purple}{9.46}} & \gold{\textcolor{purple}{2.2}} & 1       &  \silver{1.70} & \gold{\textcolor{purple}{0.37}} & \gold{\textcolor{purple}{0.16}} & 1       & 1 \\
\hyperlink{Raft-Stereo}{\textcolor{magenta}{iRaftStereo\_RVC}}~\cite{lipson2021raft,jiang2022improved}  & 3DV 2021     & 1.88 & \silver{3.03} & \bronze{2.07} & 4        &  \bronze{24.0} & \silver{13.3} & \silver{2.9} & 2       &  \bronze{1.88} & \bronze{0.55} & \silver{0.17} & 2       & 2 \\
\hyperlink{Raft-Stereo}{\textcolor{magenta}{raft+\_RVC}}~\cite{lipson2021raft}  & 3DV 2021           & \silver{1.60} & \gold{\textcolor{purple}{2.98}} & \gold{\textcolor{purple}{1.83}} & 1        & \silver{22.6} & \bronze{14.4} & \bronze{3.86} & 3       &  2.18 & 0.71 & \bronze{0.21} & 4       & 3 \\
\hyperlink{CroCo-Stereo}{\textcolor{magenta}{CroCo\_RVC}}~\cite{croco_v2}  & ICCV 2023           & 2.04 & 3.75 & 2.33 & 4        & 32.9 & 19.7 & 5.1 & 4       &  \gold{\textcolor{purple}{1.54}} & \silver{0.50} & \bronze{0.21} & 3       & 3 \\
\hyperlink{LaC}{\textcolor{magenta}{MaskLacGwcNet\_RVC}}~\cite{liu2022local_LaC}  & AAAI 2022   & \bronze{1.65} & 3.68 & 1.99 & 3        & 31.3 & 15.8 & 13.5 & 5       &  6.42 & 1.88 & 0.38 & 5       & 5 \\

\Xhline{2pt}
\multicolumn{15}{c}{\href{http://robustvision.net/rvc2020.php}{Robust Vision Challenge 2020}} \\
\hline

\hyperlink{CFNet}{\textcolor{magenta}{CFNet\_RVC}}~\cite{shen2021cfnet}  & CVPR 2021           & \silver{1.65} & \gold{3.53} & \silver{1.96} & 2        & \gold{26.2} & \gold{16.1} & 5.07 & 2       & \gold{3.7} & \gold{0.97} & \gold{0.26} & 1       & 1 \\
NLCA\_NET\_v2\_RVC \cite{rao2022rethinking} & TNNLS 2022   & \gold{\textcolor{purple}{1.51}} & \silver{3.97} & \gold{1.92} & 1        & \silver{29.4} & \silver{16.4} & 5.6 & 3        & \silver{4.11} & \silver{1.2} & \bronze{0.29} & 2       & 2 \\
HSM-Net\_RVC \cite{yang2019hierarchical_HSM} & CVPR 2019         & 2.74 & 8.73 & 3.74 & 6        & \bronze{31.2} & \bronze{16.5} & 3.44 & 1       & \bronze{4.4} & 1.51 & \silver{0.28} & 3       & 3 \\
CVANet\_RVC \cite{mehltretter2019cnn} & ICCVW 2019          & 1.76 & 4.91 & \bronze{2.28} & 3        & 58.5 & 38.5 & 8.64 & 4       & 4.58 & 1.32 & 0.32 & 4      & 4 \\
\hyperlink{AANet}{\textcolor{magenta}{AANet\_RVC}}~\cite{xu2020aanet}  & CVPR 2020           & 2.23 & 4.89 & 2.67 & 5        & 42.9 & 31.8 & 12.8 & 5       & 5.41 & 1.95 & 0.33 & 5      & 5 \\
GANetREF\_RVC \cite{zhang2019ga} & CVPR 2019        & \bronze{1.88} & \bronze{4.58} & 2.33 & 4        & 43.1 & 24.9 & 15.8 & 6       & 6.97 & \bronze{1.25} & 0.45 & 6      & 6 \\

\Xhline{2pt}
\multicolumn{15}{c}{\href{http://robustvision.net/rvc2020.php}{Robust Vision Challenge 2018}} \\
\hline

iResNet\_ROB \cite{liang2018learning_iResNet} & CVPR 2018          & \silver{2.27} & \gold{4.89} & \silver{2.71} & 2      & 45.9 & 31.7 & 6.56 & 3       & \bronze{4.67} & \silver{1.22} & \silver{0.27} & 2       & 1 \\
DN-CSS\_ROB \cite{ilg2018occlusions} & ECCV 2018          & \bronze{2.39} & \bronze{5.71} & \bronze{2.94} & 3      & \bronze{41.3} & \bronze{28.3} & 5.48 & 1       & \gold{3.0} & \gold{0.96} & \gold{0.24} & 1        & 2 \\
PSMNet\_ROB \cite{chang2018pyramid_PSMNet} & CVPR 2018           & \gold{1.79} & \silver{4.92} & \gold{2.31} & 1      & 67.3 & 47.2 & 8.78 & 6       & 5.41 & \bronze{1.31} & \bronze{0.36} & 3       & 3 \\
CBMV\_ROB \cite{batsos2018cbmv} & CVPR 2018             & 3.55 & 12.09 & 4.97 & 4     & \gold{21.6} & \gold{13.3} & 6.65 & 2       & \silver{4.66} & 2.06 & \bronze{0.36} & 4       & 4 \\
SGM\_ROB \cite{hirschmuller2007stereo} & TPAMI 2007              & 5.06 & 13.00 & 6.38 & 5     & \silver{38.6} & \silver{26.4} & 14.2 & 4       & 10.77 & 4.67 & 0.57 & 5      & 5 \\
ELAS\_ROB \cite{geiger2010efficient} & ACCV 2010             & 7.38 & 21.15 & 9.67 & 6     & 51.7 & 34.6 & 13.4 & 5       & 17.82 & 8.75 & 0.8 & 6       & 6 \\
\bottomrule

\end{tabular}
}
\caption{\textbf{Robust Vision Challenge.}}
\label{table:rvc}
\end{table*}
\begin{table*}[t]
\centering
\setlength{\tabcolsep}{8pt}
\rowcolors{2}{salmon}{white}
\scalebox{0.7}{
\begin{tabular}{ll ccc ccc ccc ccc ccc}

\Xhline{2pt}
\multirow{2}{*}{Method}  & \multirow{2}{*}{Venue} & \multicolumn{3}{c}{All} & \multicolumn{3}{c}{Class 0} & \multicolumn{3}{c}{Class 1} & \multicolumn{3}{c}{Class 2} & \multicolumn{3}{c}{Class 3} \\
\cmidrule(lr){3-5} \cmidrule(lr){6-8} \cmidrule(lr){9-11} \cmidrule(lr){12-14} \cmidrule(lr){15-17}
& &  bad2 $\downarrow$  & EPE  $\downarrow$ & Rank  & bad2$\downarrow$ & EPE  $\downarrow$ & Rank & bad2$\downarrow$ & EPE  $\downarrow$ & Rank & bad2$\downarrow$ & EPE $\downarrow$ & Rank & bad2$\downarrow$ & EPE $\downarrow$ & Rank \\

\Xhline{2pt}
\hyperlink{Raft-Stereo}{\textcolor{magenta}{RAFT-Stereo}}~\cite{lipson2021raft}        & 3DV 2021     & 35.67 & 	16.29 & 7      & 28.08 &  \silver{2.49} & 3      & 39.24 & 9.25 & 7      & 71.49 & 39.26 & 9      & 80.13 & 48.71 & 8 \\ 
\hyperlink{Raft-Stereo}{\textcolor{magenta}{RAFT-Stereo (ft)}}~\cite{lipson2021raft}     & 3DV 2021     & 34.67 &  5.26 & 6      & 30.87 &  2.70 & 5     & 	37.17 &  4.54 & 5      & 56.32 & 7.15 & 5      & 64.82 & \gold{12.32} & 1 \\ 
\hyperlink{CFNet}{\textcolor{magenta}{CFNet}}~\cite{shen2021cfnet}                   & CVPR 2021    & 61.35 & 27.61 & 8      & 60.62 & 22.05 & 9      & 56.38 & 11.64 & 9      & 74.55 & 23.44 & 7     & 81.68 & 43.51 & 7 \\
\hyperlink{CFNet}{\textcolor{magenta}{CFNet (ft)}}~\cite{shen2021cfnet}             & CVPR 2021    & 66.85 & 19.65 & 9      & 62.93 & 	9.72 & 8      & 66.37 & 11.29 & 8      & 80.66 & 38.80 & 8      & 87.28 & 42.46 & 6 \\
\hyperlink{PCVNet}{\textcolor{magenta}{PCVNet}}~\cite{zeng2023parameterized_PCVNet}           & ICCV 2023    & \gold{27.41} & \silver{6.21} & 1      & \silver{24.67} & \silver{2.49} & 2      & \silver{26.36} & \silver{3.04} & 2      & 	\gold{36.76} & \gold{3.93} & 1      & \gold{53.11} & \bronze{20.96} & 3 \\
\hyperlink{DKT-Stereo}{\textcolor{magenta}{DKT-RAFT}}~\cite{zhang2024robust_DKT}           & CVPR 2024    & 	\silver{28.60} & \bronze{6.82} & 2      & \bronze{24.81} & 2.50 & 4      & \gold{25.67} & \gold{3.03} & 1      & 	\silver{42.81} & \silver{4.77} & 2      & 	\silver{60.73} & 23.90 & 4 \\
\hyperlink{DKT-Stereo}{\textcolor{magenta}{DKT-IGEV}}~\cite{zhang2024robust_DKT}          & CVPR 2024    & 	34.48 & 9.30 & 5      & 31.01 & 5.44 & 7      & 34.90 & 3.91 & 4      & 		53.41 & 5.86 & 4      & 		\bronze{61.52} & 28.69 & 5 \\
 \hyperlink{CREStereo}{\textcolor{magenta}{CREStereo (ft)}}~\cite{li2022practical_CREStereo}           & CVPR 2022    & 	\bronze{29.53} & \gold{5.10} & 3      & \gold{24.40} & 3.09 & 6      & \bronze{30.28} & \bronze{3.31} & 3      & 			\bronze{53.14} & \bronze{4.79} & 3      & 			67.63 & \silver{18.94} & 2 \\
\hyperlink{CREStereo}{\textcolor{magenta}{CREStereo}}~\cite{li2022practical_CREStereo}          & CVPR 2022    & 	33.07 & 12.56 & 4      & 25.55 & \gold{2.12} & 1      & 36.31 & 5.62 & 6      & 				62.77 & 20.89 & 6      & 				79.50 & 53.34 & 9 \\

\bottomrule

\end{tabular}
}
\caption{\textbf{Booster Dataset.}}
\label{table:booster}
\end{table*}

\subsection{KITTI 2015}

Table \ref{table:kitti_15} collects entries from the KITTI 2015 online leaderboard concerning the stereo frameworks reviewed in our survey, with methods ranked according to the D1-all metric. The table reports three main sub-categories: 1) foundational stereo networks published in the 2020s, 2) efficient architectures published in 2020-2024, and 3) representative models from the 2010s.
Although the KITTI benchmark was already saturated before the 2020s, the most recent advances in the field allowed for further improvements, with the latest proposals from 2023-2024 establishing consistently at the very top of the leaderboard, outsitting LEAStereo \cite{yin2019hierarchical} after nearly four years.
Concerning efficient architectures, we can appreciate how the gap with state-of-the-art models has been significantly reduced, making them a viable alternative for practical applications.


\subsection{Middlebury v3}

Table \ref{table:middlebury_v3} reports the results sourced from the Middlebury v3 online benchmark, with methods being ranked according to the bad2.0 metric. We divide it into two sub-tables, grouping on top the models among those covered in this survey, and at the bottom those from previous years \cite{poggi2021synergies}. At first glance, we can appreciate a large improvement over the previous state-of-the-art, which got better and better through the five years that passed from the previous surveys. By looking at the top 6 positions of the leaderboard, we find RAFT-Stereo \cite{lipson2021raft} and five more architectures derived from it, suggesting how the former has been a game changer for a benchmark challenging as Middlebury v3. Indeed, it is worth noticing how RAFT-Stereo has been one of the very first architectures, after HSMNet, capable of processing Middlebury images at full resolution, unleashing the possibility of maintaining many more details in the final disparity maps, as well as its outstanding generalization performance contributed to its success on this benchmark. 
Selective-IGEV currently represents the state-of-the-art: on the one hand, it achieves an average error on non-occluded pixels below the pixel for the first time. On the other hand, the average error raises over 1.5 on all pixels, in particular those large ones caused by very close objects, suggesting that the occlusions may still represent an open challenge.


\subsection{Robust Vision Challenge (RVC)}

Table \ref{table:rvc} collects the leaderboards from the three editions of the Robust Vision Challenge that took place in 2018, 2020, and 2022, reported respectively from the bottom to the top. All methods are ranked on the individual benchmarks respectively according to D1-all, bad1, and avg error metrics on KITTI, Middlebury, and ETH3D, while the overall rank is computed according to the Schulze Proportional Ranking (PR) \cite{schulze2011new}.
We highlight that the challenge allowed for training on any existing dataset -- both synthetic and real ones; for a synthetic-to-real benchmark, we refer the readers to \cite{Tosi_2023_CVPR}.
The trend highlighted through the three editions is consistent with the evolution of the field we have reported in this survey.
Indeed, end-to-end deep stereo was still young in 2018, as we can notice from the bottom-most section of the table, where iResNet \cite{liang2018learning_iResNet} and PSMNet \cite{chang2018pyramid_PSMNet} are the only frameworks belonging to this category.
Then, the field developed quickly in the following two years and CFNet \cite{shen2021cfnet} ranked first in 2020, surpassing any of the networks from the 10s, with HSM-Net being the best among these latter models.
Lastly, in 2022 the first three positions were taken by models extending the optimization paradigm introduced by RAFT-Stereo \cite{lipson2021raft}, confirming how the design strategy emerged from this latter has been a game-changer in the field.
The rapid improvements observed through the three editions can be appreciated, in particular, by focusing on the error rates achieved on the Middlebury dataset, more specifically on the bad1 metric: indeed, this latter was higher than 45\% for the winner of the first edition in 2018, dropping to about 26\% for the winner of the 2020 edition and, finally, getting down to 16.5\% in the latest edition.

\subsection{Booster}

Table \ref{table:booster} reports entries from the Booster online benchmark. Specifically, we show error metrics computed at full resolution (i.e., approximately 12 Megapixels) across different categories: \textit{All}, which covers the whole regions of the images, as well as classes 0 to 3, which represents pixels belonging to materials from opaque to highly transparent/reflective. All methods are ranked, for each category, according to the bad2 score.
We can appreciate how the whole benchmark remains very difficult for current state-of-the-art stereo networks, with the very high resolution and the presence of non-Lambertian objects being the two main challenges. 
PCVNet \cite{zeng2023parameterized_PCVNet} is the absolute winner, outperforming any other framework on classes 2 and 3, while remaining competitive on 0 and 1. 
By looking at RAFT-Stereo and CREStereo, for which results are reported by using both the original weights as well as those obtained after running a fine-tuning on the Booster training set, we highlight how the error rates on the most challenges classes 2 and 3 are largely reduced after fine-tuning, proving that state-of-the-art models have the potential to learn how to deal with non-Lambertian objects from context. Nonetheless, we point out how significant efforts are still necessary to properly deal with the two, aforementioned challenges.

\section{Discussion}
\label{sec:discussion}

In this section, we highlight the key messages from our survey, highlighting the key advances made in the 2020s and suggesting potential avenues for future developments in deep stereo matching.

\textbf{Architecture Design.} As the benchmark results show, the new design strategy introduced by RAFT-Stereo has been a game changer, bringing much higher robustness to domain shifts. Most of the latest frameworks released a few months before this survey followed this new paradigm, and we expect more to come. Nonetheless, the quest for finding novel and effective architectures is not over, as witnessed by the very latest proposals \cite{guan2024neural_NMRF} achieving increasingly better results.

\textbf{Stereo Beyond RGB.} A trend that has been consolidating over the past five years is the use of other modalities, such as images from thermal, multispectral, or event cameras, as input to stereo matching networks. This brings fresh perspectives to a longstanding yet vivid field. However, online benchmarks for these new tasks are still rare, and more would help to consolidate this track.

\textbf{Open Challenges.} Despite the many successes in addressing some of the challenges anticipated by previous surveys \cite{poggi2021synergies}, some persist. Indeed, the Booster dataset \cite{Ramirez_2022_CVPR,ramirez2023booster} highlighted how images at very high resolution remain difficult to deal with, as well as non-Lambertian objects are critical, mainly due to the lack of training data or sub-optimal approaches to handle them. Similarly, challenging weather conditions still represent potential obstacles.

\textbf{Foundational Models.} Finally, in the wake of the emergence of visual foundational models for various computer vision tasks, we argue that a foundational model for stereo matching is still missing. While some attempts have been made recently for single-image depth estimation \cite{depthanything}, an effort in this direction has not yet been made for stereo.

\section{Conclusion}
\label{sec:conclusion}

In this paper, we have surveyed the advances in deep stereo matching that have emerged in the 2020s.
Building upon existing surveys \cite{poggi2021synergies,laga2020survey} whose time horizon was limited to the end of 2019, we have deeply investigated the new architectures and design trends that have appeared recently and have become the standard implementation patterns today.
We also looked at more advanced applications of stereo matching, ranging from sensor fusion to cross-spectral matching across different modalities. Finally, we have analyzed the performance of various methods on popular online benchmarks. This analysis not only identifies the current leading approaches but also sheds light on the remaining challenges and potential future research directions.
We believe that this survey can serve as a practical guide for both novices and seasoned experts, providing inspiration for their work.

\bibliographystyle{IEEEtran}
\bibliography{architectures,datasets,background_stereo, generalization,adaptation,event,other,multi-modal,tom,unsupervised}



\section*{Supplementary material}

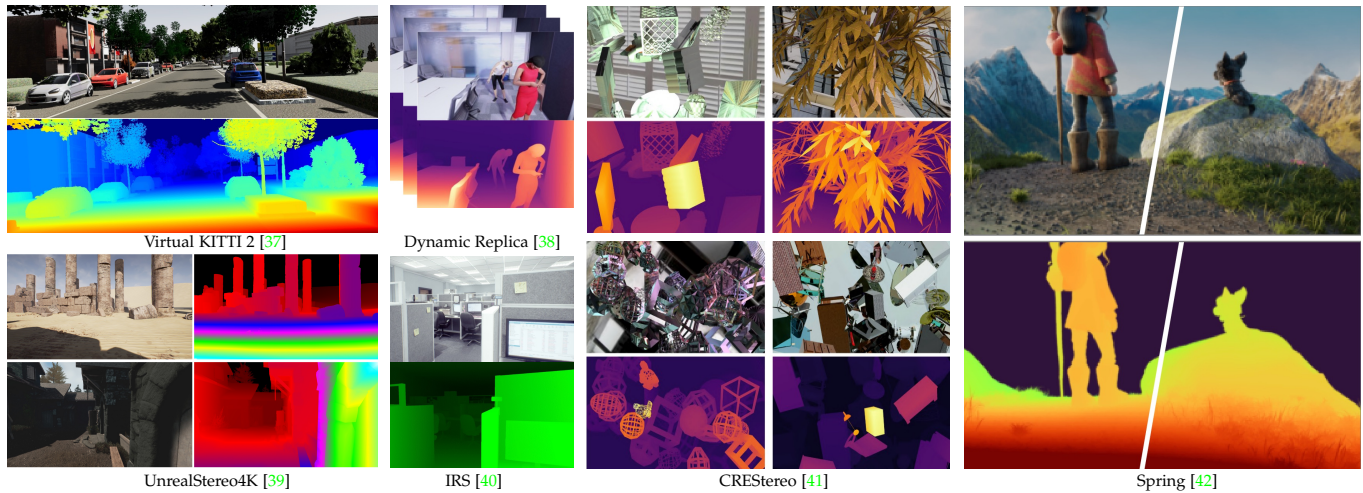

Fig. 3: **New stereo datasets in the 20s (Synthetic).**

high-quality stereo pairs, including removing vertical disparity and disparity-based filtering. However, Holopix50k contains only stereo RGB images without any ground truth disparity, making it suitable for self-supervised training of stereo networks. Moreover, due to the lack of ground truth disparity, Holopix50k is not typically used as a benchmark for evaluating stereo vision algorithms.

**InStereo2K** [3][4]. The InStereo2K dataset is a large-scale stereo dataset designed for indoor scenes, containing 2050 pairs of stereo images with highly accurate disparity maps obtained using a structured light system. The dataset is split into a training set of 2000 image pairs and a test set of 50 image pairs, with a resolution of $1080 \times 860$ pixels. InStereo2K covers a wide range of indoor scenes, including offices, classrooms, bedrooms, living rooms, and dormitories. The structured light system used to generate the dataset consists of two color cameras with a resolution of $1280 \times 960$ pixels and a projector with a resolution of $1024 \times 768$ pixels.

### 2.1.2 Multimodal Stereo Datasets

**CATS** [45][5]. The Color and Thermal Stereo (CATS) dataset is a large-scale benchmark consisting of approximately 1400 images captured by a sensor platform with two visible-band cameras at a resolution of $1280 \times 960$ pixels, two long wave infrared (LWIR) thermal cameras at a resolution of $640 \times 480$ pixels, and a LiDAR for high-accuracy (¡2mm) ground truth. The dataset features 100 indoor and 80 outdoor cluttered scenes with diverse objects under various lighting and environmental conditions. CATS provides rectified stereo pairs in color, thermal, and cross-modality configurations, along with corresponding ground truth disparity maps generated by projecting the LiDAR data onto the images using a novel semi-automatic calibration method.

**MVSEC** [46][6]. The Multi Vehicle Stereo Event Camera dataset (MVSEC) collects data from both indoor and outdoor environments with event cameras. Specifically, a 10cm-baseline stereo camera made with DAVIS 346B sensors collects event frames at 346×260 resolution, together with a grayscale stereo camera mounting Aptina MT9V034 sensors working at $752 \times 480$ resolution, a Velodyne Puck LITE (16-lines) to collect sparse depth data, an IMU, motion capture, and GPS. This sensor suite allows for extracting accurate ground truth pose and disparity maps – the latter obtained by accumulating LiDAR point clouds according to a LiDAR odometry pipeline (LOAM [47]). Sequences are collected in a mixture of indoor and outdoor environments from different vehicles: 6 from a hexacopter, 5 from a car, one from a motorcycle, and 2 when the camera is handheld, both during day and night.

**DSEC** [7][7]. The DSEC (Driving Stereo Event Camera) dataset is designed to address the challenges faced by standard cameras in adverse lighting conditions such as night, sunrise and sunset. It provides a rich set of sensory data, including a wide-baseline stereo setup with two color frame cameras ($1440 \times 1080$ resolution, 51 cm baseline) and two high-resolution monochrome event cameras ($640 \times 480$ resolution, 60 cm baseline), which offer unique advantages like high temporal resolution, high dynamic range, reduced motion blur, and low latency. The dataset contains 53 sequences collected in various illumination conditions and environments in Switzerland, totaling over 4 TB of raw data, and also includes LiDAR data, RTK GPS measurements, accurate time synchronization, and calibration between sensors. DSEC provides ground truth disparity and incorporates sequences with challenging illumination conditions.

**M3ED** [5][8]. This dataset provides 57 indoor/outdoor scenes collected with a compact multi-sensor block mounted on three different vehicles – a car, a UAV, and a quadruped robot. The event stereo camera mounts Prophesee Gen 4 sensors, capturing frames at $1280 \times 720$ resolution with a 12cm baseline, accompanied by a color stereo camera mounting OVC 3b sensors working at $1280 \times 800$ resolution with an equivalent baseline. A 64-line Ouster OS1-64U LiDAR is used to generate semi-dense ground-truth disparity maps, by accumulating single scans through LiDAR odometry (FasterLIO [48]). In total, M3ED provides about 3TB of raw

4. https://github.com/YuhuaXu/StereoDataset
5. https://bigdatavision.org/CAT/download.html
6. https://daniilidis-group.github.io/mvsec/
7. https://dsec.ifi.uzh.ch/
8. https://m3ed.io/

and processed data.

**Gated Stereo** [8][9]. Gated Stereo is a long-range dataset that was captured to facilitate training and evaluation of depth estimation methods that leverage gated imaging and stereo cues. Gated imaging enables robust long-range depth measurement by using active illumination and measuring time-of-flight, overcoming limitations of passive techniques. The dataset was acquired by driving over 1000 km in Germany with a sensor suite comprising a long-range LiDAR, an RGB stereo camera, and a synchronized NIR gated stereo camera. The gated camera captures $1280 \times 720$ resolution images at 120 Hz, providing 3 gated slices and 2 ambient exposures enabled by active VCSEL illumination. The RGB stereo camera provides $1920 \times 1080$ images at 30 Hz. The dataset contains over 107,000 samples captured across day, night, and varying weather conditions. After being curated for scenario diversity, it is divided into 54,320 samples for training, 728 for validation, and 2,463 for testing.

**RGB-MS** [49][10]. The RGB-MS dataset tackles the novel challenge of registering synchronized RGB and multi-spectral (MS) images with vastly different resolutions via stereo matching. It comprises 34 indoor RGB-MS image pairs from 13 scenes, with 12.4MP RGB images paired with 2.2MP 10-band MS images and annotated with high-resolution semi-dense ground truth disparities. These were generated using an additional 12.2MP RGB camera during acquisition - static scenes perturbed with patterns enabled robust disparity estimation from active RGB stereo pairs via cost volume integration, semi-global matching, outlier removal, and warping to the RGB-MS views. 11K unlabeled RGB-RGB-MS triplets are also provided in both indoor and outdoor environments.

**MS²** [1][11]. The Multi-Spectral Stereo (**MS²**) dataset is a large-scale collection of synchronized multi-modal data captured for autonomous driving, comprising around 195K data pairs of stereo RGB images ($2448 \times 2048$ resolution, 15fps), stereo near-infrared (NIR) images ($1280 \times 720$ resolution, 15fps), stereo long-wave infrared (thermal) images ($640 \times 512$ resolution, 30fps), stereo LiDAR point clouds (10fps), and GNSS/IMU information. The data was acquired across diverse environments like cities, residential areas, roads, and campuses, under varying conditions including daytime, nighttime, cloudy, and rainy weather. The thermal stereo images are provided along with semi-dense ground truth depth maps generated by accumulating successive LiDAR sweeps and utilizing interpolated odometry.

**RPS** [6][12]. The Real Polarimetric Stereo dataset (RPS) is a collection of 1,300 stereo pairs – 1,000 of which have been collected indoors, while the remaining 300 are in outdoor environments. Two Lucid PHX050S-Q polarization cameras have been used to capture stereo polarization images at 1280×960 resolution, paired with an RGB-D camera for collecting ground-truth depth. Each polarization camera in the pair captures four polarization images, with different polarizer angles in a single shot.

## 2.2 Synthetic

We now introduce more datasets obtained through graphic engines, allowing for generating vast amounts of stereo images and dense ground truth disparities with little effort. Fig. 3 shows an overview of same samples taken from these datasets.

**HR-VS** [50][13]. The High-res Virtual Stereo (HR-VS) dataset, presented in [50] along with the HSMNet architecture, is a synthetic high-resolution stereo dataset generated using the Carla simulator [51]. It contains 780 pairs of stereo images at $2056 \times 2464$ pixel resolution collected under 4 weather conditions in Town01, with a camera baseline of 0.54m. The dataset mimics real-world driving scenarios by limiting the disparity range to [9.66, 768] pixels and the depth range to [2.52, 200] meters, making it suitable for training high-resolution stereo matching algorithms for autonomous driving and urban scene understanding.

**Virtual KITTI 2** [37][14] Building upon its predecessor, Virtual KITTI [52], this dataset offers photo-realistic renderings of driving scenarios generated using the Unity game engine. The dataset comprises 5 image sequences, each a clone of a real-world sequence from the KITTI tracking benchmark, with a resolution of $1242 \times 375$ pixels. What sets Virtual KITTI 2 apart is its diverse set of renditions for each sequence, including variations in weather conditions like fog and rain, as well as modifications to the camera configuration, such as rotations of $15°$ and $30°$. In addition to RGB images, the dataset provides ground truth data for depth, optical flow, scene flow, instance segmentation, and semantic segmentation.

**TartanAir** [53][15]. The TartanAir dataset is a large-scale, challenging dataset designed for robot navigation tasks, with a special focus on stereo vision and SLAM-related problems. It provides synchronized stereo RGB images across 1037 long motion sequences in 30 diverse environments, covering structured urban and indoor scenes as well as unstructured natural environments, resulting in over 1 million frames. Collected in photo-realistic simulation using the Unreal Engine and AirSim plugin, the dataset also includes depth images, segmentation labels, camera poses, occupancy grid maps, optical flow, stereo disparity, and simulated LiDAR measurements. TartanAir covers a wide range of motion patterns and includes challenging scenes with dynamic lighting, low illumination, adverse weather, and dynamic objects, aiming to bridge the gap between simulation and real-world performance.

**UnrealStereo4K** [39] [16]. The UnrealStereo4K dataset is a large-scale synthetic stereo dataset created using the Unreal Engine and the open-source plugin UnrealCV. It consists of 8 scenes, including both indoor and outdoor environments, with a total of 8,000 stereo pairs at $3840 \times 2160$px resolution. The dataset is divided into 7,720 training pairs, 80 validation pairs, and 200 in-domain test pairs. An additional 200 out-of-domain test pairs from an unseen scene are provided to evaluate the generalization ability of stereo matching

---

9. https://light.princeton.edu/gatedstereo/
10. https://cvlab-unibo.github.io/rgb-ms-web/
11. https://sites.google.com/view/multi-spectral-stereo-dataset
12. https://github.com/Ethereal-Tian/DPS-Net
13. https://github.com/gengshan-y/high-res-stereo?tab=readme-ov-file
14. https://europe.naverlabs.com/Research/Computer-Vision/Proxy-Virtual-Worlds/
15. https://theairlab.org/tartanair-dataset/
16. https://github.com/fabiotosi92/SMD-Nets

methods. The baseline is set to 20cm for indoor environments and 50cm for outdoor environments. The camera poses are randomly sampled, and invalid camera poses and stereo pairs are filtered out based on minimum depth, image intensity, and disparity map gradient criteria to ensure non-trivial geometry structure in the dataset.

**IRS** [40][17]. The IRS (Indoor Robotics Stereo) dataset is a large-scale synthetic stereo dataset designed for indoor robotics applications, containing over 100,000 stereo image pairs (84,946 for training and 15,079 for testing) with a resolution of $960 \times 540$ pixels. The dataset covers four indoor scene types (home, office, restaurant, and store) and 70 different scene layouts, with more than 2,091 identical furniture objects of various types placed within the constructed spaces. IRS provides high-quality and dense ground truth labels for surface normal and disparity maps, which are crucial for training deep learning models. The dataset is generated using a customized version of Unreal Engine 4 with originally implemented plug-ins, ensuring that the rendered images closely resemble real-world scenes in terms of material texture, light reflection and transmission, shadows, bloom, and lens flare.

**CREStereo** [41][18]. In the paper introducing the CREStereo architecture, a synthetic dataset for training stereo matching algorithms is also proposed. This dataset, generated using Blender, includes left-right image pairs and corresponding dense disparity maps. The dataset scenes feature a variety of object shapes sourced from ShapeNet and custom models with complex characteristics such as holes and open-work structures. Additionally, the dataset incorporates complex lighting environments with random colors and luminance, using real-world images as textures for objects and backgrounds. To ensure a wide range of disparity, objects are randomly positioned within the virtual cameras' field of view, with varying distances and scales.

**SimStereo** [54][19]. The Active-Passive SimStereo dataset contains 515 computer-generated image pairs with a resolution of $640 \times 480$ pixels, rendered using a physically-based rendering engine. This ensures that the images closely resemble real-world scenes in terms of material texture, light reflection, transmission, shadows, bloom, and lens flare. The dataset is split into a training set of 412 image pairs and a test set of 103 image pairs, covering various shapes, depth ranges, and styles. The test set includes realistic scenes and abstract compositions to effectively benchmark the performance of stereo vision models. SimStereo provides both active and passive frames for each scene, allowing for a direct comparison of the relative performance of algorithms in both scenarios.

**Spring** [42][20]. The Spring dataset introduces a large, high-resolution, high-detail, synthetic benchmark for scene flow, optical flow, and stereo estimation. Based on rendered scenes from the open-source Blender movie "Spring", it provides 6,000 stereo image pairs from 47 sequences at $1920 \times 1080$ resolution, along with 12,000 ground truth frames for stereo and 23,812 ground truth frames for motion - 60x more than KITTI 2015 and 15x more than MPI Sintel. To adequately assess fine details, the dataset uses a novel evaluation methodology based on super-resolved ($3840\times2160$px) ground truth, and provides focused evaluations on different image regions including high-detail, unmatched, non-rigid, sky, and large-displacement areas. The dataset is accompanied by a public benchmark website to upload, analyze and compare novel methods.

**Dynamic Replica** [38][21]. Dynamic Replica is a synthetic stereo video dataset featuring 524 videos of animated humans and animals in everyday scenes, designed for training temporally consistent models. The dataset includes 484 training videos, 20 validation videos (both 10 seconds long with 300 frames each), and 20 test videos (30 seconds long), all rendered at $1280 \times 720$ resolution. The dataset is based on Facebook Replica reconstructions, using 375 3D human scans and 13 animal categories, with randomized camera baselines between 4cm and 30cm in the training subset. All samples contain ground-truth depth maps, optical flow, segmentation masks, and camera parameters for both stereo views.

### 2.3 Evaluation Metrics

The evaluation of stereo methods is typically performed by comparing the estimated disparity maps against ground truth data using several metrics. A widely adopted metric is the End-Point Error (EPE)↓, defined as:

$$\text{EPE} = \frac{1}{N} \sum_p |D_p - D_p^{gt}| \quad (3)$$

where $D_p$ and $D_p^{gt}$ are the predicted and ground truth disparity values for pixel $p$, respectively, and $N$ is the total number of pixels. The EPE calculates the average absolute difference between the predicted and ground truth disparity values across all pixels. Optionally, the Root Mean Squared Error (RMSE) is also taken into account in some benchmarks, such as Middlebury v3 and Booster:

$$\text{RMSE} = \sqrt{\frac{1}{N} \sum_p (|) D_p - D_p^{gt})^2} \quad (4)$$

Moreover, existing benchmarks commonly report the bad-$\tau$ error↓, defined as the percentage of pixels whose disparity estimation deviates by more than $\tau$ pixels from the ground truth:

$$\text{bad-}\tau = \frac{1}{N} \sum_p \delta(|D_p - D_p^{gt}| > \tau) \quad (5)$$

where $\delta$ is the indicator function that returns 1 if the condition is true, and 0 otherwise. On the KITTI 2015 benchmark, a variant of the bad metric, the D1 error, is adopted:

$$D1 = \frac{1}{N} \sum_p \delta(|D_p - D_p^{gt}| > 3 \wedge \\ |D_p - D_p^{gt}| > 0.05 \cdot D_p^{gt}) \quad (6)$$

considering as outliers those disparities whose error surpasses 3 pixels and 5% of the ground-truth value.

---

17. https://github.com/HKBU-HPML/IRS
18. https://github.com/megvii-research/CREStereo
19. https://ieee-dataport.org/open-access/active-passive-simstereo
20. http://spring-benchmark.org
21. https://dynamic-stereo.github.io/



\end{document}